\documentclass{article}

\usepackage[final,nonatbib]{neurips_2021}

\usepackage[utf8]{inputenc} % allow utf-8 input
\usepackage[T1]{fontenc}    % use 8-bit T1 fonts
\usepackage{hyperref}       % hyperlinks
\usepackage{url}            % simple URL typesetting
\usepackage{booktabs}       % professional-quality tables
\usepackage{amsfonts}       % blackboard math symbols
\usepackage{nicefrac}       % compact symbols for 1/2, etc.
\usepackage{microtype}      % microtypography
\usepackage{xcolor}         % colors

\usepackage{graphicx}
\usepackage{subfig}
\usepackage{bm}
\usepackage{amssymb}
\usepackage{multicol}
\usepackage{amsmath}
\usepackage{amsthm}
\newtheorem{theorem}{Theorem}
\newtheorem{lemma}{Lemma}
\newtheorem{remark}{Remark}
\usepackage[ruled]{algorithm2e}
\usepackage{wrapfig}
\usepackage{pifont}
\usepackage{url}

\title{Believe What You See: Implicit Constraint Approach for Offline Multi-Agent Reinforcement Learning}

\author{%
	Yiqin Yang$^1$\footnotemark[2] , Xiaoteng Ma$^1$\footnotemark[2] \footnotemark[3], Chenghao Li$^1$, Zewu Zheng$^1$, \\
	\textbf{Qiyuan Zhang$^2$, Gao Huang$^1$, Jun Yang$^1$\footnotemark[3] ,  Qianchuan Zhao$^1$} \\
    $^1$Tsinghua University, $^2$Harbin Institute of Technology\\
    \{yangyiqi19, ma-xt17, lich18\}@mails.tsinghua.edu.cn, zzheng17@126.com, \\ zhangqiyuan19@hit.edu.cn, \{gaohuang, yangjun603, zhaoqc\}@tsinghua.edu.cn
}

\begin{document}
	\maketitle
	\renewcommand{\thefootnote}{\fnsymbol{footnote}}
	\footnotetext[2]{Equal Contribution.}
	\footnotetext[3]{Equal Corresponding.}
	
	\begin{abstract}
	Learning from datasets without interaction with environments~(Offline Learning) is an essential step to apply Reinforcement Learning (RL) algorithms in real-world scenarios.
	However, compared with the \emph{single-agent} counterpart, offline \emph{multi-agent} RL introduces more agents with the larger state and action space, which is more challenging but attracts little attention.
	We demonstrate current offline RL algorithms are ineffective in multi-agent systems due to the accumulated extrapolation error. 
    In this paper, we propose a novel offline RL algorithm, named \emph{Implicit Constraint Q-learning} (ICQ), which effectively alleviates the extrapolation error by only trusting the state-action pairs given in the dataset for value estimation.
    Moreover, we extend ICQ to multi-agent tasks by decomposing the joint-policy under the implicit constraint.
    Experimental results demonstrate that the extrapolation error is successfully controlled within a reasonable range and insensitive to the number of agents.
	We further show that ICQ achieves the state-of-the-art performance in the challenging multi-agent offline tasks~(StarCraft II).
	Our code is public online at \href{https://github.com/YiqinYang/ICQ}{https://github.com/YiqinYang/ICQ}.
	\end{abstract}
	
	\section{Introduction} \label{introduction}
	Recently, reinforcement learning~(RL), an active learning process, has achieved massive success in various domains ranging from strategy games~\cite{ye2020mastering} to recommendation systems~\cite{cao2020adversarial}.
	However, applying RL to real-world scenarios poses practical challenges: interaction with the real world, such as autonomous driving, is usually expensive or risky.
	To solve these issues, offline RL is an excellent choice to deal with practical problems~\cite{agarwal2020optimistic, levine2020offline, peng2019advantage, siegel2020keep, fujimoto2019benchmarking, nachum2019dualdice, ajay2020opal, laroche2019safe, vuong2018supervised, fakoor2020p3o}, aiming at learning from a fixed dataset without interaction with environments.
	
	The greatest obstacle of offline RL is the distribution shift issue~\cite{fujimoto2019off}, which leads to extrapolation error, a phenomenon in which unseen state-action pairs are erroneously estimated.
	Unlike the online setting, the inaccurate estimated values of unseen pairs cannot be corrected by interacting with the environment.
	Therefore, most off-policy RL algorithms fail in the offline tasks due to intractable over-generalization.
	Modern offline methods~(e.g., Batch-Constrained deep Q-learning~(BCQ)~\cite{fujimoto2019off}) aim to enforce the learned policy to be close to the behavior policy or suppress the $Q$-value directly.
	These methods have achieved massive success in challenging single-agent offline tasks like D4RL~\cite{fu2020d4rl}.
	
	However, many decision processes in real-world scenarios belong to multi-agent systems, such as intelligent transportation systems~\cite{adler2002cooperative}, sensor networks~\cite{rabbat2004distributed}, and power grids~\cite{callaway2010achieving}.
	%To mitigate the accumulated error in multi-agent systems, we analyze the inducement of the extrapolation error theoretically in single-agent.
	%Meanwhile, we analyze its changes when extending to the multi-agent setting.
% 	We demonstrate that
% 	\emph{unseen state-action pairs will grow exponentially as the number of agents increases in multi-agent systems, accumulating the extrapolation error quickly}.
    Compared with the single-agent counterpart, the multi-agent system has a much larger action space, which grows exponentially with the increasing of the agent number. When coming into the offline scenario, \emph{the unseen state-action pairs will grow exponentially as the number of agents increases, accumulating the extrapolation error quickly}.
	The current offline algorithms are unsuccessful in multi-agent tasks even though they adopt the modern value-decomposition structure~\cite{ma2021modeling, sunehag2017value, li2021celebrating}.
	As shown in Figure~\ref{MMDP}, our results indicate that BCQ, a state-of-the-art offline algorithm, has divergent $Q$-estimates in a simple multi-agent MDP environment~(e.g., BCQ (4 agents)). The extrapolation error for value estimation is accumulated quickly as the number of agents increases, significantly impairing the performance.
	
	Based on these analyses, we	propose the Implicit Constraint Q-learning~(ICQ) algorithm, which effectively alleviates the extrapolation error as no unseen pairs are involved in estimating $Q$-value.
	Motivated by an implicit constraint optimization problem, ICQ adopts a SARSA-like approach~\cite{sutton2018reinforcement} to evaluate $Q$-values and then converts the policy learning into a supervised regression problem.
% 	Besides, ICQ adopts an implicit constraint optimization method, which converts the policy learning into a supervised regression problem and meanwhile guarantees the convergence property of $Q$-function theoretically.
	By decomposing the joint-policy under the implicit constraint, we extend ICQ to the multi-agent tasks successfully.
	To the best of our knowledge, our work is the first study analyzing and addressing the extrapolation error in multi-agent reinforcement learning.
	
	We evaluate our algorithm on the challenging multi-agent offline tasks based on StarCraft II~\cite{samvelyan2019starcraft}, where a large number of agents cooperatively complete a task.
	Experimental results show that ICQ can control the extrapolation error within a reasonable range under any number of agents and learn from complex multi-agent datasets.
	Further, we evaluate the single-agent version of ICQ in D4RL, a standard single-agent offline benchmark. 
	The results demonstrate the generality of ICQ for a wide range of task scenarios, from single-agent to multi-agent, from discrete to continuous control.
	
	\begin{figure}[t]
		\centering
		\subfloat[BCQ.]
		{\label{BCQ trajectory}\includegraphics[width=2.41in]{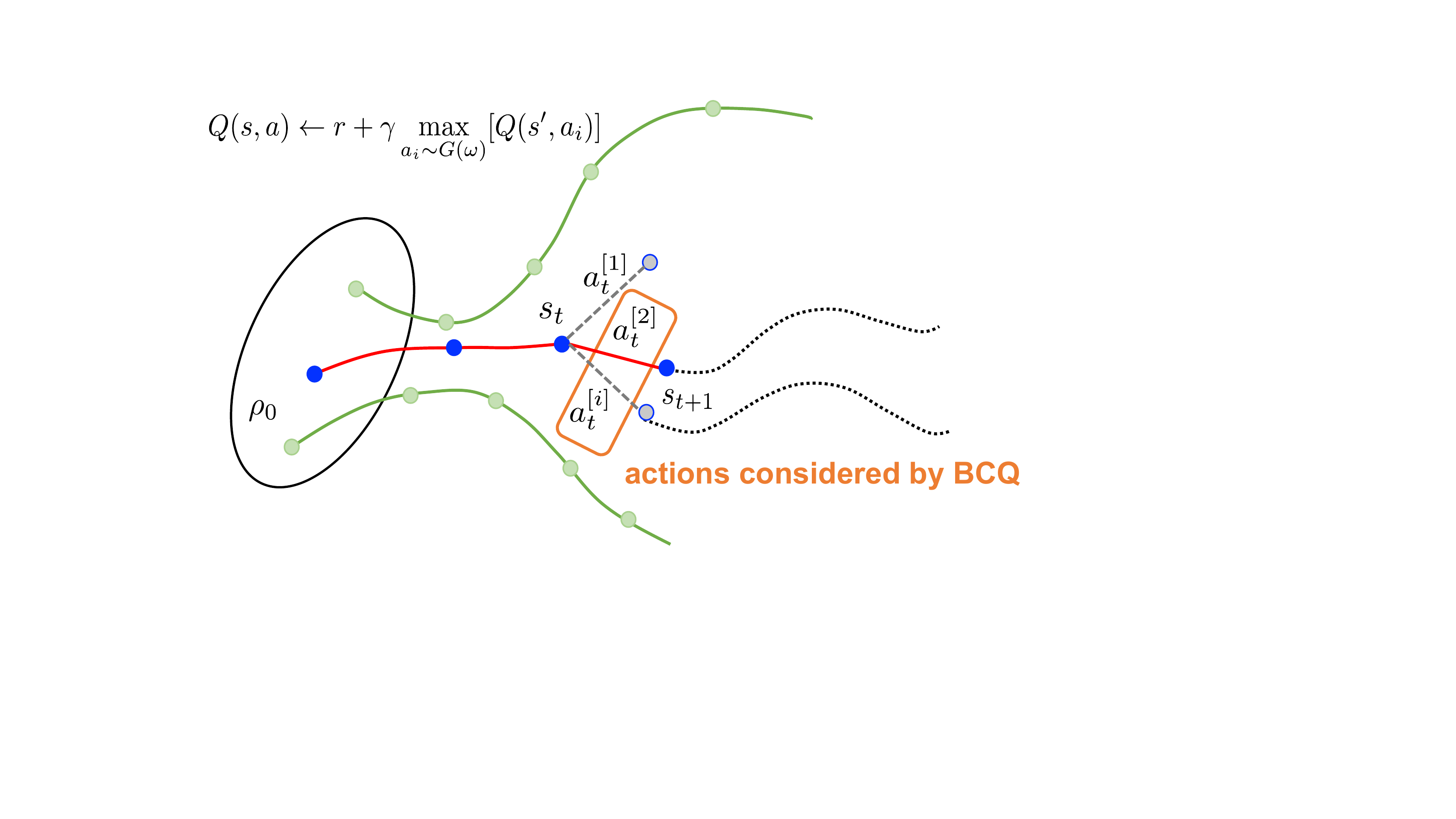}}
		\hspace{5mm}
		\subfloat[ICQ.]
		{\label{ICQ trajectory}\includegraphics[width=2.86in]{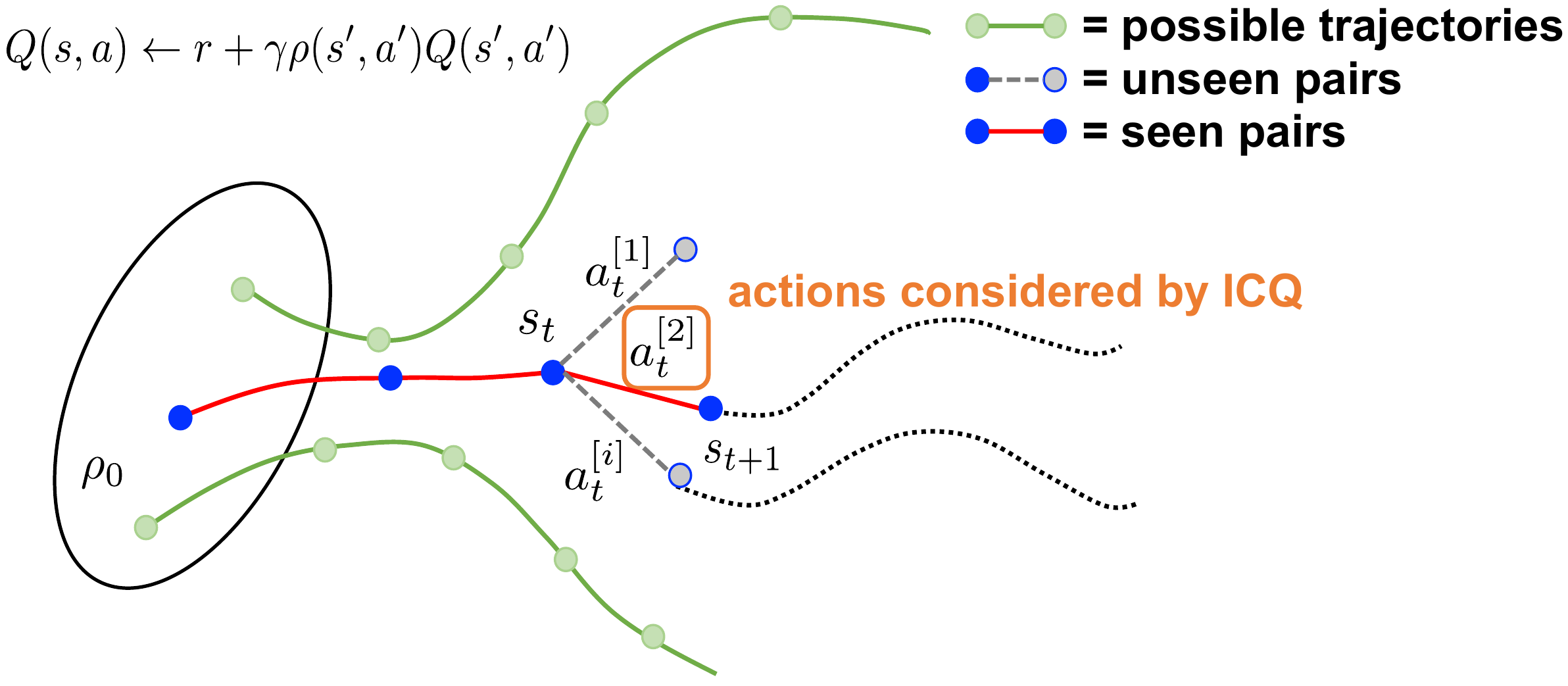}}
		\caption{The comparison between ICQ and BCQ for the target $Q$-value estimation.
		The spots denote states, and the connections between spots indicate actions.
		The red solid-lines denote seen pairs, and the gray dotted-lines are unseen pairs.
		(a) BCQ estimates $Q$-value in a defined similar action set~(orange) while unseen pairs still exist in the set with low probability.
		(b) ICQ only adopts seen pairs~(orange) in the training set for $Q$-value estimation.
		}
		\label{Trajecotries}
	\end{figure}

	\section{Background}
	\label{background}
	\textbf{Notation.} The fully cooperative multi-agent tasks are usually modeled as the Dec-POMDP~\cite{oliehoek2016concise} consisting of the tuple $G =\left < S, A, P, r, \Omega, O, n, \gamma \right>$.
	Let $s \in S$ denote the true state of the environment.
	At each time step $t \in \mathbb{Z}^+$, each agent $i \in N \equiv \{1,\dots,n\}$ chooses an action $a^i \in A$, forming a joint action $\bm{a} \in \mathbf{A} \equiv A^n$.
	Let $P(s'\mid s, \bm{a}):S \times \mathbf{A} \times S \to [0, 1]$ denote the state transition function. 
	All agents share the same reward function $r(s, \bm{a}):S \times \mathbf{A}\to \mathbb{R}$.
	
	We consider a partially observable scenario in which each agent draws individual observations $o^i \in \Omega$ according to the observation function $O(s,a): S \times \mathbf{A} \to \Omega$.
	Each agent has an action-observation history $\tau^i\in \mathbf{T} \equiv (\Omega \times \mathbf{A})^t$, on which it conditions a stochastic policy $\pi^i(a^i\mid \tau^i)$ parameterized by $\theta_i : \mathbf{T} \times \mathbf{A} \to [0, 1]$.
	The joint action-value function is defined as $Q^{\bm{\pi}}(\bm{\tau}, \bm{a}) \triangleq \mathbb{E}_{s_{0:\infty}, \bm{a}_{0:\infty}}\left[\sum_{t=0}^{\infty}\gamma^t r_t \mid s_0=s, \bm{a}_0=\bm{a}, \bm{\pi} \right]$, where $\bm{\pi}$ is the joint-policy with parameters $\theta=\left<\theta_1,\dots, \theta_n \right>$.
	Let $\mathcal{B}$ denote the offline dataset, which contains trajectories of the behavior policy $\bm{\mu}$.

	We adopt the \textit{centralized training and decentralized execution}~(CTDE) paradigm~\cite{son2019qtran}.
	During training, the algorithm has access to the true state $s$ and every agent's action-observation history $\tau_i$, as well as the freedom to share all information between agents. However, during execution, each agent has access only to its action-observation history.
	
	\textbf{Batch-constrained deep Q-learning}~(BCQ) is a state-of-the-art offline RL method, which aims to avoid selecting an unfamiliar action at the next state during a value update.
	Specifically, BCQ optimizes $\pi$ by introducing perturbation model $\xi(\tau, a, \Phi)$ and generative model $G(\tau; \varphi)$ as follows
	\begin{equation}
	\begin{aligned}
	\pi(\tau) = \mathop{\arg\max}_{a^{[i]}+\xi(\tau,a^{[i]}, \Phi)}& Q^{\pi}(\tau, a^{[i]}+\xi(\tau, a^{[i]}, \Phi); \phi), {\quad \rm s.t. \quad} \{a^{[i]} \sim G(\tau; \varphi)\}_{i=1}^{m},
	\end{aligned}
	\end{equation}
	where $\pi$ selects the highest valued action from a collection of $m$ actions sampled from the generative model $G(\tau; \varphi)$, which aims to produce only previously seen actions.
% 	\begin{wrapfigure}{h}{0.23\textwidth}
% 	    \vspace{-1em}
% 		\includegraphics[width=0.23\textwidth]{fig/BCQ_perturbation.pdf}
% 		\caption{Mechanism of BCQ.}
% 	    \vspace{-1em}
% 		\label{BCQ_perturbation}
% 	\end{wrapfigure}
	The perturbation model $\xi(\tau, a^{[i]}, \Phi)$ is adopted to adjust action $a^{[i]}$ in the range $[-\Phi, \Phi]$ to increase the diversity of actions.
% 	The working mechanism of BCQ is shown in Figure~\ref{BCQ_perturbation}.
	
	%The generative model $G(\tau; \varphi)$ is trained to match the state-action pairs sampled from $\mathcal{D}$ by minimizing the cross-entropy loss function~\cite{de2005tutorial}.
	%Let $\{a_t^{[i]}\}_{i=0}^m$ denote the action set.
	%As for the discrete version of BCQ, if $\frac{a^{[i]}\sim G(\tau;\psi)}{\max\{a^{[i]} \sim G(\tau; \varphi)\}_{i=1}^{m} } \leq \zeta$, $a^{[i]}$ is considered an unfamiliar action and $\xi(\tau, a^{[i]})$ will mask $a^{[i]}$ in the maximizing the $Q$-value operation.
	%The action-value function is updated by minimizing the following loss
	%\begin{equation}
	%\begin{aligned}
	%L(\phi, \psi) = \mathbb{E}_{\mu}\left[\left(r + \gamma \max_{a^{[i]}+\xi(\tau, a^{[i]})} Q^{\pi}(\tau', 
	%a^{[i]}+\xi(\tau, a^{[i]});\phi',\psi') - Q^{\pi}(\tau, a; \phi, \psi)
	%\right)^2 \right].
	%\end{aligned}
	%\end{equation}

	\section{Analysis of Accumulated Extrapolation Error in Multi-Agent RL}
	\label{motivation}
	In this section, we theoretically analyze the extrapolation error propagation in offline RL, which lays the basis for Section~\ref{Method}.
	The extrapolation error mainly attributes the out-of-distribution~(OOD) actions in the evaluation of $Q^{\pi}$~\cite{fujimoto2019off,kumar2019stabilizing}.
	To quantify the effect of OOD actions, we define the state-action pairs within the dataset as \emph{seen} pairs.
	Otherwise, we name them as \emph{unseen} pairs.
	We demonstrate that the extrapolation error propagation from the unseen pairs to the seen pairs is related to the size of the action space, which grows exponentially with the increasing number of agents. We further design a toy example to illustrate the inefficiency of current offline methods in multi-agent tasks. 
% 	We demonstrate that the extrapolation error is gradually accumulated as the number of agents increases, which lays the basis for Section~\ref{Method}.
% 	Our theoretical result indicates that the key of offline RL is to reduce the effect of unseen pairs when evaluating the seen ones.
    
    \subsection{Extrapolation Error Propagation in Offline RL}
    Following the analysis in BCQ~\cite{fujimoto2019off}, we define the tabular estimation error\footnote{Note that we adopt a different definition of extrapolation error with BCQ. The $\epsilon_{\rm MDP}(\tau, a)$ is regraded as the extrapolation error in BCQ, while the generalization error of unseen pairs $\epsilon_{\rm EXT}(\tau, a)$ is considered in this work.} as $\epsilon_{\rm MDP}(\tau, a) \triangleq Q_M^{\pi}(\tau, a) - Q^{\pi}_{\mathcal{B}}(\tau, a)$ (here we abuse $\tau$ to denote the state for analytical clarity), where the $M$ denotes the true MDP and $\mathcal{B}$ denotes a new MDP computed from the batch by $P_{\mathcal{B}}(\tau^\prime \mid \tau, a) = \mathcal{N}(\tau,a,\tau^\prime) / \sum_{\tilde{\tau}}\mathcal{N}(\tau,a,\tilde{\tau})$. BCQ~\cite{fujimoto2019off} has shown that $\epsilon_{\rm MDP}(\tau, a)$ has a Bellman-like form with the extrapolation error $\epsilon_{\rm EXT}(\tau, a)$ as the "reward function":
    \begin{equation}
    \begin{aligned}
        \epsilon_{\rm MDP}(\tau, a) &\triangleq \epsilon_{\rm EXT}(\tau, a) + \sum_{\tau^\prime} P_M(\tau^\prime \mid \tau, a)\gamma \sum_{a^\prime}\pi(a^\prime \mid s^\prime)\epsilon_{\rm MDP}(\tau^\prime, a^\prime), \\
        % \text{where } 
        \epsilon_{\rm EXP}(\tau, a) &= \sum_{\tau^\prime}\big(P_M(\tau^\prime \mid \tau,a) - P_{\mathcal{B}}(\tau^\prime \mid \tau,a)\big) \Big(r(\tau,a,\tau^\prime) + \gamma\sum_{a^\prime}\pi(a^\prime \mid \tau^\prime)Q^{\pi}_{\mathcal{B}}(\tau^\prime, a^\prime)\Big).    
    \end{aligned}
    \end{equation}
    For the seen state-action pairs, $\epsilon_{\rm EXT}(\tau, a)=0$ since $P_M(\tau^\prime \mid \tau,a) - P_{\mathcal{B}}(\tau^\prime \mid \tau,a)=0$ in the deterministic environment. In contrast, the $\epsilon_{\rm EXT}(\tau, a)$ of unseen pairs is uncontrollable and depends entirely on the initial values in tabular setting or the network generalization in DRL.
	
	To further analyze how the extrapolation error in the unseen pairs impacts the estimation of actions in the dataset, we partition $\bm{\epsilon_{\rm MDP}}$ and $\bm{\epsilon_{\rm EXT}}$ as $\bm{\epsilon_{\rm MDP}}= [\bm{\epsilon_{\rm s}}, \bm{\epsilon_{\rm u}]^{\rm T}}$ and $ \bm{\epsilon_{\rm EXT}} = [\bm{0}, \bm{\epsilon_{\rm b}}]^{\rm T}$ respectively according to seen and unseen state-action pairs. 
% 	As interaction with the environment is not allowed, we can only optimize $\epsilon_{\rm s}$ while not eliminating $\epsilon_{\rm u}$.
	Let denote the transition matrix of the state-action pairs as $P^{\pi}_M(\tau^\prime, a^\prime\mid \tau, a) = P_{M}(\tau^\prime \mid \tau, a) \pi(a' \mid \tau^\prime)$.
	We decompose the transition matrix as $P^{\pi}_M = \left[P^{\pi}_{\rm s,s}, P^{\pi}_{\rm s,u};P^{\pi}_{\rm u,s}, P^{\pi}_{\rm u,u}
	\right]$ according to state-action pairs' property (e.g., $P^{\pi}_{\rm s,u}(\tau^\prime_{\rm u}, a_{\rm u}' \mid \tau_{\rm s}, a_{\rm s}) = P_{M}(\tau^\prime_{\rm u} \mid \tau_{\rm s}, a_{\rm s}) \pi(a_{\rm u}' \mid \tau_{\rm u}')$ denotes the transition probability from seen to unseen pairs). Then the extrapolation error propagation can be described by the following linear system:
	\begin{equation}
	\begin{aligned}
	\left[\begin{matrix}
	\bm{\epsilon_{\rm s}} \\
	\bm{\epsilon_{\rm u}}
	\end{matrix}\right] = \gamma  \left[
	\begin{matrix}
	P^{\pi}_{\rm s, s} & P^{\pi}_{\rm s, u} \\
	P^{\pi}_{\rm u, s} & P^{\pi}_{\rm u, u}
	\end{matrix}
	\right] \left[\begin{matrix}
	\bm{\epsilon_{\rm s}} \\
	\bm{\epsilon_{\rm u}}
	\end{matrix}\right] + 
     \left[\begin{matrix}
	\bm{0} \\
	\bm{\epsilon_{\rm b}}
	\end{matrix}\right]
	\end{aligned}.
	\end{equation}

Based on the above definitions, we have the following conclusion.
	\begin{theorem}
		\label{Theorem: epsilon}
		Given a deterministic MDP, the propagation of $\bm{\epsilon_{\rm b}}$ to $\bm{\epsilon_{\rm s}}$ is proportional to $\|P^{\pi}_{\rm s,u}\|_\infty$:
	\begin{equation}
	    \left\| \bm{\epsilon_{\rm s}} \right\|_\infty \leq \frac{\gamma \left\| P^{\pi}_{\rm s, u}\right\|_\infty  }{(1 - \gamma)\left(1 - \gamma \left\|  P^{\pi}_{\rm s, s} \right\|_\infty \right)}\left\| \bm{\epsilon_{\rm b}} \right\|_\infty.
	\end{equation}
	\end{theorem}
	
	The above theorem indicates the effect of extrapolation error on seen state-action pairs is directly proportional to $\|P^{\pi}_{\rm s,u}\|_\infty$. In the practice, $\|P^{\pi}_{\rm s,u}\|_\infty$ is related to the size of action space and the dataset. If the action space is enormous, such as a multi-agent task with a number of agents, we need a larger amount of data to reduce $\|P^{\pi}_{\rm s,u}\|_\infty$. However, the dataset size in offline learning tasks is generally limited. Moreover, when using the networks to approximate the value function, $\bm{\epsilon_{\rm b}}$ does not remain constant as $Q_\mathcal{B}(\tau_{\rm u}, a_{\rm u})$ could be arbitrary during training, making the $Q$-values extreme large even for the seen pairs. 
	For these reasons, we have to enforce the $P^{\pi}_{\rm s,u} \to 0$ by avoiding using OOD actions. 
% 	Researchers are interested in $P^{\pi}_{\rm s,u}$ as we can change it by optimizing $\pi(a_{\rm u}' \mid s_{\rm u}')$
% 	while it is hard to quantify $P^{\pi}_{\rm u,u}$ as we do not know $P_{M}(s'_{\rm u} \mid s_{\rm u}, a_{\rm u})$.
	For example, BCQ utilizes an auxiliary generative model to constrain the target actions within a familiar action set (see Section~\ref{background} for a detailed description). However, the error propagation heavily depends on the accuracy of the generative model and is intolerable with the agent number increasing. We will demonstrate this effect in the following toy example.

	\begin{figure}[t]
	\centering
	\subfloat[Two-state MMDP.]
	{\label{MMDP_1}\includegraphics[width=1.8in]{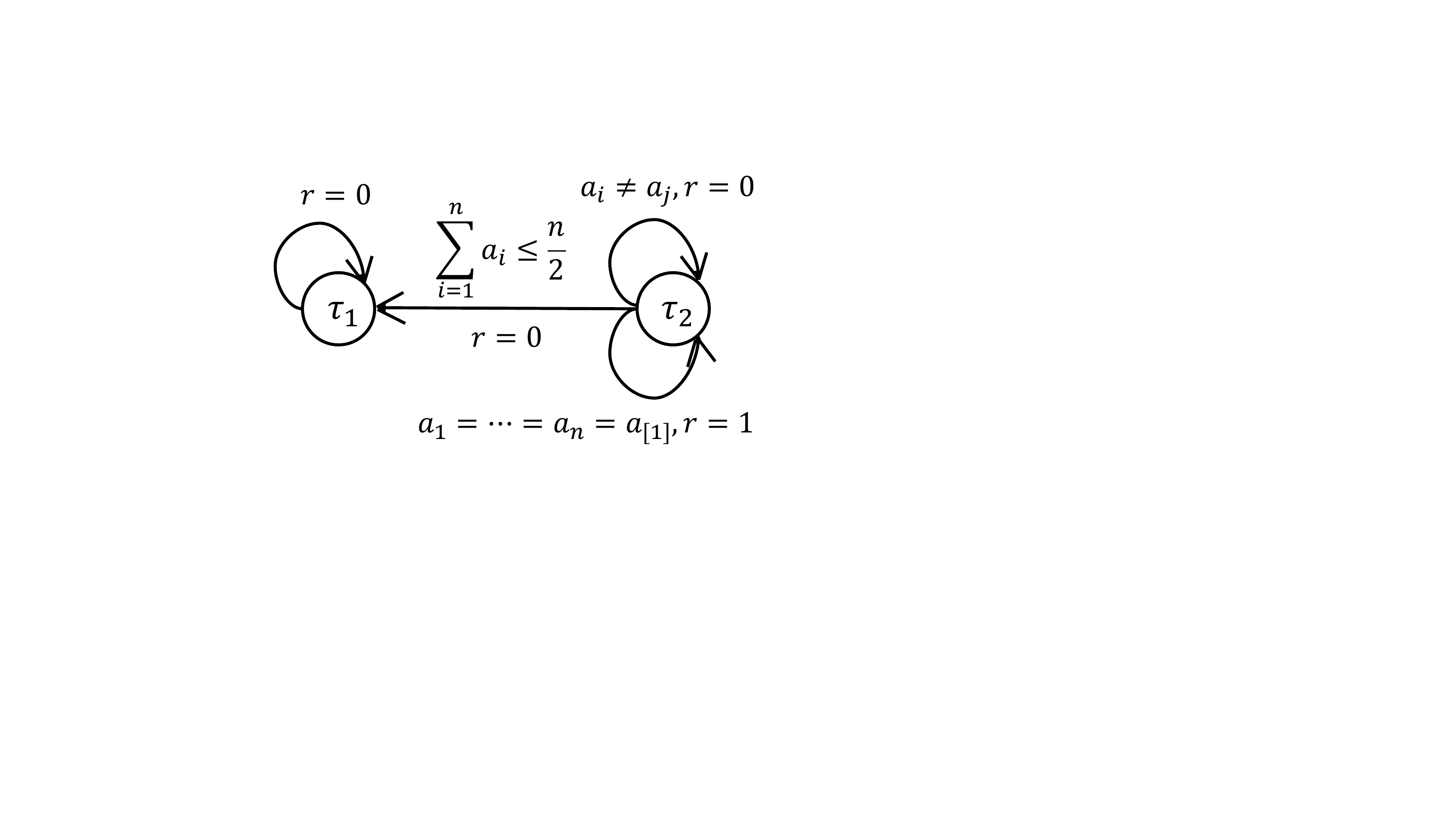}}
	\hspace{5mm}
	\subfloat[Estimated value of the joint action-value function.]
	{\label{MMDP_2}\includegraphics[width=3.0in]{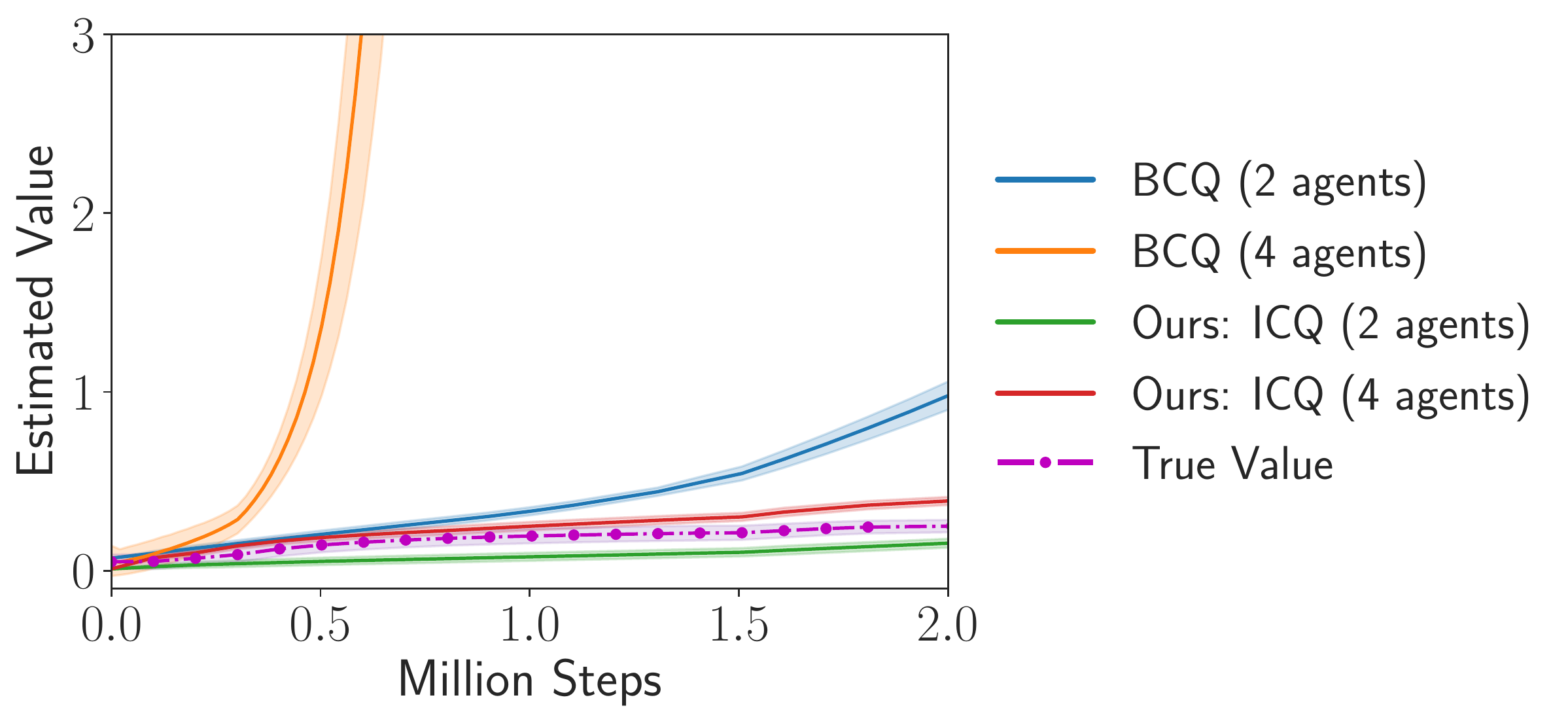}}
	\caption{(a) An MMDP where $Q$-estimates of BCQ will diverge as the number of agents increases.
	(b) The learning curve of the joint action-value function while running several agents in the given MMDP.
	The true values are similar in this task with different agent numbers, calculated by averaging the Monte-Carlo estimation under different agents.
	The $Q$-estimates of BCQ~(4 agents) diverge while our algorithm~(ICQ) has accurate $Q$-estimates.
    Please refer to Appendix~\ref{MMDP addtional results} for the complete results.
	}
	\label{MMDP}
	\end{figure}

	\subsection{Toy Example}
	\label{Toy Example}
	We design a toy two states Multi-Agent Markov Decision Process~(MMDP) to illustrate the accumulated extrapolation error in multi-agent tasks~(see Figure~\ref{MMDP_1}).
	All agents start at state $\tau_2$ and explore rewards for 100 environment steps by taking actions $a_{[1]}=0$ or $a_{[2]}=1$.
	The optimal policy is that all agents select $a_{[1]}$. 
	The MMDP task has sparse rewards.
	The reward is 1 when following the optimal policy, otherwise, the reward is 0.
	The state $\tau_2$ will transfer to $\tau_1$ if the joint policy satisfies $\sum_{i=1}^n a_i\leq \frac{n}{2}$ at $\tau_2$, while the state $\tau_1$ will never return to $\tau_2$.
	
	We run BCQ and our method ICQ on a limited dataset, which only contain 32 trajectories generated by QMIX. Obviously, the number of unseen state-action pairs exponentially grows as the number of agents increases.
	We control the amount of valuable trajectories ($r=1$) in different datasets equal for fair comparisons. 
	The multi-agent version of BCQ shares the same value-decomposition structure as ICQ~(see Appendix~\ref{Baselines Details}).
	
	As shown in Figure~\ref{MMDP_2}, the joint action-value function learned by BCQ gradually diverges as the number of agents increases while ICQ maintains a reasonable $Q$-value.
	The experimental result is consistent with Theorem~\ref{Theorem: epsilon}, and we provide an additional analysis for the toy example in Appendix~\ref{Proof of Theorem 1}.
	In summary, we show theoretically and empirically that the extrapolation error is accumulated quickly as the number of agents increases and makes the $Q$-estimates easier to diverge.

	% 2.35 2.85
	\section{Implicit Constraint Approach for Offline Multi-Agent RL}
	\label{Method}
	In this section, we give an effective method to solve the accumulated extrapolation error in offline Multi-Agent RL based on the analysis of Section~\ref{motivation}.
	From the implementation perspective, we find that a practical approach towards offline RL is to estimate target $Q$-value without sampled actions from the policy in training.
	We propose Implicit Constraint Q-learning~(ICQ), which only trusts the seen state-action pairs in datasets for value estimation. 
	Further, we extend ICQ to multi-agent tasks with a value decomposition framework and utilize a $\lambda$-return method to balance the variance and bias.

    \subsection{The Implicit Constraint Q-learning~(ICQ) Approach}
	\label{ICQ framework}
% 	\textbf{Comparison between Offline and Off-policy RL.}
	Based on the analysis of Section~\ref{motivation}, we find that the extrapolation error can be effectively alleviated by enforcing the actions within the dataset when calculating the target values, which is the most significant difference between offline and off-policy RL.
	For a formal comparison of off-policy and offline algorithms, we first introduce the standard Bellman operator $\mathcal{T}^{\pi}$ as follows:
	\begin{equation}
	\begin{aligned}
	(\mathcal{T}^{\pi}Q)(\tau, a) \triangleq Q(\tau, a) + \mathbb{E}_{\tau^\prime}[r + \gamma \mathbb{E}_{a'\sim \textcolor{red}{\pi}}[Q(\tau', a')] - Q(\tau, a)]
	\end{aligned}.
	\end{equation}
	Many off-policy evaluation methods, such as the Tree Backup~\cite{ernst2005tree} and Expected SARSA~\cite{santamaria1997experiments}, are designed based on this operator.
	However, when coming into the offline setting, the standard Bellman operator suffers from the OOD issue as the actions  sampled from current policy $\pi$ are adopted for target $Q$-value estimation.
	A natural way to avoid the OOD issue is adopting the importance sampling measure~\cite{neal2001annealed}:
	\begin{equation}
	\begin{aligned}
	(\mathcal{T}^{\pi}Q)(\tau, a) = Q(\tau, a) + \mathbb{E}_{\tau^\prime}[r + \gamma \mathbb{E}_{a'\sim \textcolor{red}{\mu}}[\rho(\tau', a') Q(\tau', a')] - Q(\tau, a)]
	\end{aligned},
	\end{equation}
	
	where $\rho(\tau', a') \triangleq \frac{\pi(a'\mid\tau')}{\mu(a'\mid \tau')}$ denotes the importance sampling weight.
	If we can calculate $\rho(\tau', a')$ \emph{with action $a'$ sampled from $\mu$ rather than $\pi$}, the unseen pairs will be avoided for target $Q$-value estimation. In this case, the extrapolation error is theoretically avoided since $P^{\pi}_{\rm s,u} \to 0$.
    The estimated $Q$-value based on the above operation would be stable even in complex tasks with enormous action space.
	However, in most real-world scenarios, it is hard to obtain the exact behavior policy to calculate $\rho(\tau', a')$, e.g., using expert demonstrations.
	Fortunately, we find that the solution of following implicit constraint optimization problem is efficient to compute the desired importance sampling weight.
	% To solve these issues, ICQ adopts the implicit constraint optimization method to construct the relationship between $\pi$ and $\mu$.
	
	\subsubsection{Implicit Constraint Q-learning} In offline tasks, the policies similar to the behavior policy are preferred while maximizing the accumulated reward $Q^{\pi}(\tau, a)$, i.e., $D_{\rm KL}(\pi \parallel \mu)[\tau] \leq \epsilon$. The policy optimization with the behavior regularized constraint can be described in the following problem:
	\begin{equation}
	\label{optimization problem}
	\begin{aligned}
	\pi_{k+1} = \mathop{\arg\max}_{\pi}\mathbb{E}_{a\sim\pi(\cdot\mid\tau)}[Q^{\pi_k}(\tau, a)], \quad{\rm s.t. \quad}  D_{\rm KL}(\pi \parallel \mu)[\tau] \leq \epsilon.
	\end{aligned}
	\end{equation}
	This problem has well studied in many previous works~\cite{peters2010relative,abdolmaleki2018maximum,wu2019behavior}. Note that the objective is a linear function of the decision variables $\pi$ and all constraints are convex functions. Thus we can obtain the optimal policy $\pi^*$ related to $\mu$ through the KKT condition~\cite{dreves2011solution}, for which the proof is in Appendix~\ref{Proof of Theorem 1.5}:
	\begin{equation}
	\label{optimal pi}
	\begin{aligned}
	\pi^*_{k+1}(a\mid\tau) = \frac{1}{Z(\tau)}\mu(a\mid\tau)\exp\left(\frac{Q^{\pi_k}(\tau, a)}{\alpha}\right),
	\end{aligned}
	\end{equation}
	where $\alpha > 0$ is the Lagrangian coefficient and $Z(\tau) = \sum_{\tilde{a}} \mu(\tilde{a}\mid\tau)\exp\left(\frac{1}{\alpha}Q^{\pi_k}(\tau, \tilde{a})\right)$ is the normalizing partition function.
	% which is calculated approximately by softmax operation over mini-batch samples
	Next, we calculate the ratio between $\pi$ and $\mu$ by relocating $\mu$ to the left-hand side:
	\begin{equation}
	\begin{aligned}
	\label{optimal policy}
	\rho(\tau, a) = \frac{\pi^*_{k+1}(a\mid\tau)}{\mu(a\mid\tau)} = \frac{1}{Z(\tau)}\exp\left(\frac{Q^{\pi_k}(\tau, a)}{\alpha}\right).
	\end{aligned}
	\end{equation}
	
	Motivated on Equation~\ref{optimal policy}, we define the Implicit Constraint Q-learning operator as
    \begin{equation}
        \begin{aligned}
        \label{ICQ_operator}
        \mathcal{T}_{\rm ICQ}Q(\tau,a)=r + \gamma \mathbb{E}_{a'\sim \mu }\left[\frac{1}{Z(\tau^\prime)} \exp\left(\frac{ Q\left(\tau', a'\right)} {\alpha}\right) Q\left(\tau', a'\right)\right]
        \end{aligned}.
    \end{equation}
    Thus we obtain a SARAR-like algorithm which not uses any unseen pairs.
    
    \textbf{Comparison with previous methods.} While BCQ learns an action generator to filter unseen pairs in $Q$-value estimation, it cannot work in enormous action space due to the error of the generator~(see Figure~\ref{Trajecotries}). Instead, in the value update of ICQ, we do not use the sampled actions to compute the target values, thus we alleviate extrapolation error effectively. There are some previous works, such as AWAC~\cite{nair2020accelerating} and AWR~\cite{peng2019advantage}, addressing the offline problem with similar constrained problem in Equation~\ref{optimization problem}. 
    However, these methods only impose the constraint on the policy loss and adopt the standard Bellman operator to evaluate $Q$-function, which involves the unseen actions or converges to the value of behavior policy $\mu$.
    Differently, we re-weight the target $Q(\tau', a')$ with the importance sampling weight derived from the optimization problem, which makes the estimated value closer to the optimal value function.

	\subsubsection{Theoretical Analysis} 
    The ICQ operator in Equation~\ref{ICQ_operator} results in a SARSA-like algorithm, which be re-written as:
    \begin{equation}
        \begin{aligned}
        \mathcal{T}_{\rm ICQ}Q(\tau,a)=r + \gamma \sum_{a'\in\mathcal{B}}\left[\frac{1}{Z(\tau^\prime)} \mu(a^\prime \mid \tau^\prime)\exp\left(\frac{1}{\alpha}Q\left(\tau', a'\right)\right) Q\left(\tau', a'\right)\right]
        \end{aligned}.
    \end{equation}
    This update rule can be viewed as a regularized softmax operator~\cite{song2019revisiting,pan2020softmax} in the offline setting.
    When $\alpha \to \infty$, $\mathcal{T}_{\rm ICQ}$ approaches $\mathcal{T}^\mu$. When $\alpha \to 0$, $\mathcal{T}_{\rm ICQ}$ becomes the batch-constrained Bellman optimal operator $\mathcal{T}_{\rm BCQ} $~\cite{fujimoto2019off}, which constrains the possible actions with respect to the batch:
    \begin{equation}
        \begin{aligned}
        \mathcal{T}_{\rm BCQ} Q(\tau, a) = r + \gamma \max_{a'\in\mathcal{B}}Q(\tau', a').
        \end{aligned}
    \end{equation}
    $\mathcal{T}_{\rm BCQ}$ has been shown to converge to the optimal action-value function $Q^*$ of the batch, which means $\lim_{k\to\infty}\mathcal{T}_{\rm BCQ}^k Q_0=Q^*$ for arbitrary $Q_0$.
    Based on this result, we show that iteratively applying $\mathcal{T}_{\rm ICQ}$ will result in a $Q$-function not far away from $Q^*$:
    \begin{theorem}
	\label{ICQ_convergence}
	    Let $\mathcal{T}^k_{\rm ICQ} Q_0$ denote that the operator $\mathcal{T}_{\rm ICQ}$ are iteratively applied over an initial action-value function $Q_0$ for $k$ times. Then, we have 
	    $\forall (\tau,a)$, $\mathop{\lim\sup}_{k\to \infty}\mathcal{T}_{\rm ICQ}^{k} Q_0(\tau,a) \leq Q^*(\tau,a)$,
	    \begin{equation}
	        \begin{aligned}
	        \label{ICQ_upper_bound}
	        \mathop{\lim\inf}_{k\to \infty}\mathcal{T}_{\rm ICQ}^{k}Q_0(\tau,a) \ge Q^*(\tau,a) - \frac{\gamma(|A_{\tau}|-1)}{(1-\gamma)}\max\left\{\frac{1}{(\alpha^{-1}+1)C+1}, \frac{2Q_{\rm max}}{1+C\exp(\alpha^{-1})}\right\},
	        \end{aligned}
	    \end{equation}
	    where $|A_{\tau}|$ is the number of seen actions for state $\tau$, $C \triangleq\inf_{\tau\in S}\inf_{2\leq i\leq |A_{\tau}|}\frac{\mu(a_{[1]} \mid \tau)}{\mu(a_{[i]}\mid \tau)}$ and $\mu(a_{[1]} \mid \tau)$ denotes the probability of choosing the expert action according to behavioral policy $\mu$.
	    Moreover, the upper bound of $\mathcal{T}^k_{\rm BCQ} Q_0$ - $\mathcal{T}^k_{\rm ICQ} Q_0$ decays exponentially fast as a function of $\alpha$.
	\end{theorem}
	
% 	\begin{wrapfigure}{h}{0.25\textwidth}
% 	    \vspace{-1em}
% 		\includegraphics[width=0.23\textwidth]{fig/Projecting.pdf}
% 		\caption{Projection-based policy optimization.}
% 	    \vspace{-2em}
% 		\label{Projecting}
% 	\end{wrapfigure}
	While $\mathcal{T}_{\rm ICQ}$ is not a contraction~\cite{asadi2017alternative} (similar with the softmax operator), the $Q$-values are still within a reasonable range.
    Further, $\mathcal{T}_{\rm ICQ}$ converges to $\mathcal{T}_{\rm BCQ}$ with an exponential rate in terms of $\alpha$. 
    Our result also quantifies the difficulty in offline RL problems.
    Based on the definition of $\mu(a_{[i]}| \tau)$, $C$ shows the proportion of the expert experience in the dataset.
    A larger $C$ corresponds to more expert experience, which induces a smaller distance between $\mathcal{T}_{\rm ICQ}^k Q_0(\tau,a)$ and $Q^*(\tau,a)$. In contrast, with a small $C$, the expert experience is few and the conservatism in learning is necessary. 
    
	\subsubsection{Algorithm}	
	Based on the derived operator $\mathcal{T}_{\rm ICQ}$ in Equation~\ref{optimal policy}, we can learn $Q(\tau, a;\phi)$ by minimizing 
	\begin{equation}
	\begin{aligned}
	\label{ICQ_Q_loss}
	\mathcal{J}_Q(\phi) = \mathbb{E}_{\tau, a,\tau^\prime, a^\prime \sim \mathcal{B}}\left[r + \gamma \frac{1}{Z(\tau')}\exp\left(\frac{Q\left(\tau', a'; \phi'\right)}{\alpha}\right) Q\left(\tau', a'; \phi'\right) - Q\left(\tau, a; \phi\right)\right]^2,
	\end{aligned}
	\end{equation}
	where the $Q$-network and the target $Q$-network are parameterized by $\phi$ and $\phi^\prime$ respectively.
	
	As for the policy training, we project the non-parametric optimal policy $\pi_{k+1}^*$ in Equation~\ref{optimal pi} into the parameterized policy space $\theta$ by minimizing the following KL distance, which is implemented on the data distribution of the batch:
	\begin{equation}
	\begin{aligned}
	\label{policy_update}
	\mathcal{J}_{\pi}(\theta) & =  \mathbb{E}_{\tau \sim \mathcal{B}}\left[D_{\rm KL}\left(\pi^*_{k+1} \| \pi_{\theta}\right)[\tau] \right] = \mathbb{E}_{\tau \sim \mathcal{B}}\left[-\sum_a\pi^*_{k+1}(a\mid\tau)\log\frac{\pi_{\theta}(a\mid\tau)}{\pi^*_{k+1}(a\mid\tau)} \right] \\
	&\overset{(a)}{=} \mathbb{E}_{\tau \sim \mathcal{B}}\left[\sum_a \frac{\pi^*_{k+1}(a\mid\tau)}{\mu(a\mid\tau)}\mu(a\mid\tau)\left(-\log\pi_{\theta}(a\mid\tau)\right) \right] \\
	& \overset{(b)}{=} \mathbb{E}_{\tau, a \sim \mathcal{B}}\left[-\frac{1}{Z(\tau)}\log(\pi(a\mid\tau;\theta))\exp\left(\frac{Q(\tau, a)}{\alpha} \right)\right],
	\end{aligned}
	\end{equation}
	% $d_{\mu}(\tau)$ denotes data distribution under $\mu$
	where $(a)$ ignores $\mathbb{E}_{\tau \sim \mathcal{B}}\left[\sum_a \pi^*_{k+1}(a\mid\tau)\log\pi^*_{k+1}(a\mid\tau)\right]$ that is not related to $\theta$, and $(b)$ applies the importance sampling weight derived in Equation~\ref{optimal policy} under forward KL constraint. 
    Note that tuning the $\alpha$ parameter in Equation~\ref{policy_update} between 0 and $\infty$ interpolates between $Q$-learning and behavioral cloning.
	See Appendix~\ref{algorithm appendix} for the complete workflow of the ICQ algorithm.
    We provide two implementation options to compute the normalizing partition function $Z(\tau)$, which is discussed in detail in Appendix~\ref{ICQ Details}.
    
	\subsection{Extending ICQ to Multi-Agent Tasks}
	\label{ICQ Multi-Agent}
	In the previous section, we propose an implicit constraint $Q$-learning framework by re-weighting target $Q$-value $Q(\tau',a')$ in the critic loss, which is efficient to alleviate the extrapolation error.
	We next extend ICQ to multi-agent tasks.
	For notational clarity, we name the \textbf{M}ulti- \textbf{A}gent version of ICQ as ICQ-MA.
	% The complete algorithm is shown in Appendix~\ref{algorithm appendix}.

	\subsubsection{Decomposed Multi-Agent Joint-Policy under Implicit Constraint}
	\begin{wrapfigure}{r}{5cm}
		\includegraphics[width=2.0in]{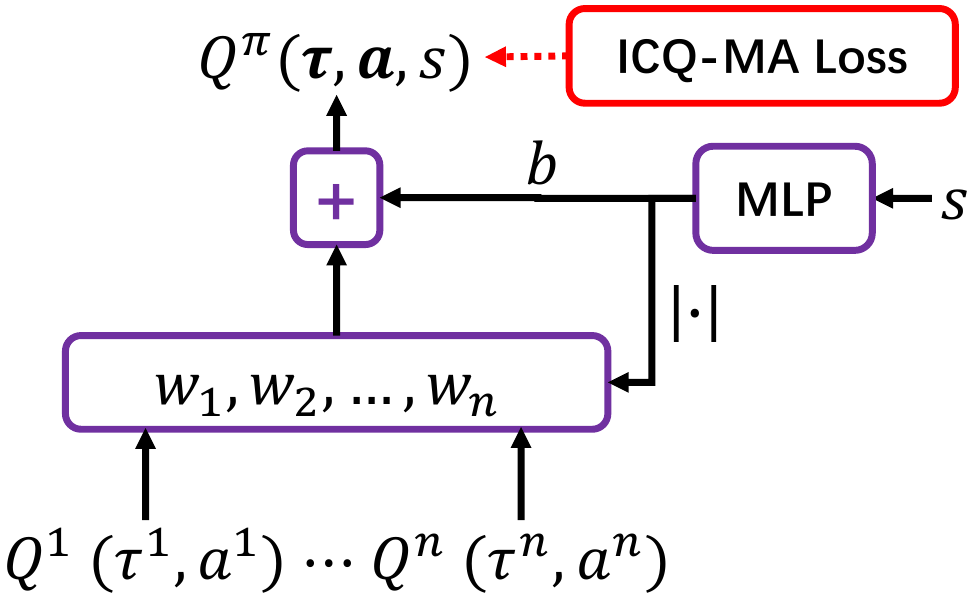}
		\caption{Mixer Network.}
		\vspace{-1em}
		\label{Mix Network}
	\end{wrapfigure}
	Under the CTDE framework, we have to train individual policies for decentralized execution.
	Besides, it is also challenging to compute
	$\mathbb{E}_{\bm{\mu}}[\rho(\bm{\tau}', \bm{a}') Q^{\bm{\pi}}(\bm{\tau}', \bm{a}')]$ in multi-agent policy evaluation as its computational complexity is $O(|A|^n)$.
	To address the above issues, we first define the joint-policy as $\bm{\pi}(\bm{a}\mid \bm{\tau})\triangleq\Pi_{i\in N}\pi^i(a^i\mid \tau^i)$, and then introduce a mild value-decomposition assumption:
	\begin{equation}
	Q^{\bm{\pi}}(\bm{\tau}, \bm{a}) = \sum_{i}w^i(\bm{\tau})Q^i(\tau^i, a^i) + b(\bm{\tau}),
	\end{equation}
	where $w^i(\bm{\tau})\geq0$ and $b(\bm{\tau})$ are generated by the Mixer Network whose inputs are global observation-action history~(see Figure~\ref{Mix Network}).
	Based on the above assumptions, we propose the decomposed multi-agent joint-policy under implicit constraint in the following theorem:
	\begin{theorem}
		Assuming the joint action-value function is linearly decomposed, we can decompose the multi-agent joint-policy under implicit constraint as follows
		\begin{equation}
		\begin{aligned}
		\bm{\pi} = \mathop{\arg\max}_{\pi^1,\dots,\pi^n} \sum_i \mathbb{E}_{\tau^i, a^i\sim \mathcal{B}}\left[\frac{1}{Z^i(\tau^i)}
		\log (\pi^i(a^i\mid \tau^i)) \exp\left(\frac{w^i(\bm{\tau})Q^i(\tau^i, a^i)}{\alpha}\right)\right]
		\end{aligned},
		\end{equation}
		where $Z^i(\tau^i) = \sum_{\tilde{a}^i}\mu^i(\tilde{a}^i \mid \tau^i) \exp\left(\frac{1}{\alpha}w^i(\bm{\tau})Q^i(\tau^i, \tilde{a}^i)\right)$ is the normalizing partition function.
	\end{theorem}
	%Please refer to Appendix~\ref{Proof of Theorem 3} for the detailed proof.
	
	The decomposed multi-agent joint-policy has a concise form.
	We can train individual policies $\pi^i$ by minimizing
	\begin{equation}
	    \begin{aligned}
	    \mathcal{J}_{\bm{\pi}}(\theta) = \sum_i \mathbb{E}_{\tau^i, a^i\sim \mathcal{B}}\left[-\frac{1}{Z^i(\tau^i)}\log(\pi^i(a^i\mid\tau^i;\theta_i))\exp\left(\frac{w^i(\bm{\tau})Q^{i}(\tau^i, a^i)}{\alpha} \right)\right].
	    \end{aligned}
	\end{equation}
	Besides, $w^i(\bm{\tau})$ achieves the trade-off between the roles of agents.
	If some agents have important roles, the value of corresponding $w^i(\bm{\tau})$ is relatively large.
	Also, if $w^i(\bm{\tau}) \to 0$, $\pi^i$ is approximately considered as the behavior cloning policy.
	As for the policy evaluation, we train $Q(\bm{\tau},\bm{a};\phi, \psi)$ by minimizing 
	\begin{equation}
	    \begin{aligned}
	    \mathcal{J}_Q(\phi, \psi)=\mathbb{E}_{\mathcal{B}}\left[\sum_{t\geq 0}(\gamma\lambda)^t\left(	r_t + \gamma \frac{1}{Z(\bm{\tau}_{t+1})}	\exp\left(\frac{Q(\bm{\tau}_{t+1},\bm{a}_{t+1})}{\alpha}\right)	Q(\bm{\tau}_{t+1},\bm{a}_{t+1})-Q(\bm{\tau}_t, \bm{a}_t)\right)\right]^2,
	    \end{aligned}
	\end{equation}
	where $Q(\bm{\tau}_{t+1},\bm{a}_{t+1}) =\sum_iw^i(\bm{\tau}_{t+1};\psi')Q^i(\tau^i_{t+1}, a^i_{t+1};\phi'_i)-b(\bm{\tau}_{t+1};\psi')$.
	
		\subsubsection{Multi-Agent Value Estimation with \texorpdfstring{$\lambda$}{lambda}-return}
	As the offline dataset contains complete behavior trajectories, it is natural to accelerate the convergence of ICQ with the $n$-step method. 
	Here we adopt $Q(\lambda)$~\cite{munos2016q} to improve the estimation of ICQ, which weights the future temporal difference signal with a decay sequence $\lambda^t$.
% 	Based on the analysis in Section~\ref{ICQ framework}, we can effectively alleviate $\epsilon_{\rm s}$ in single-agent tasks by optimizing $P^{\pi}_{\rm s,u} \to 0$.
% 	However, the generalization error of the neural network in seen pairs will lead to large variance with the number of agents increasing, making learning unstable~\cite{wang2020off}.
	Further, the constraint in Equation~\ref{optimization problem} implicitly meets the convergence condition of $Q(\lambda)$.
    Therefore, we extend the ICQ operator in Equation~\ref{ICQ_operator} to $n$-step estimation, which is similar to $Q(\lambda)$:
	\begin{equation}
	\label{multi-agent policy evaluation}
	\begin{aligned}
	(\mathcal{T}^{\lambda}_{\rm ICQ}Q)(\bm{\tau}, \bm{a}) \triangleq Q(\bm{\tau}, \bm{a}) + \mathbb{E}_{\mu}\left[\sum_{t\geq 0}(\gamma\lambda)^t\left(
	r_t + \gamma\rho(\bm{\tau}_{t+1}, \bm{a}_{t+1})Q(\bm{\tau}_{t+1},\bm{a}_{t+1})-Q(\bm{\tau}_t, \bm{a}_t)\right) \right],
	\end{aligned}
	\end{equation}
	where $\rho(\bm{\tau}_{t}, \bm{a}_{t})=\frac{1}{Z(\bm{\tau}_t)}\exp(\frac{1}{\alpha}Q(\bm{\tau}_t, \bm{a}_t))$ and hyper-parameter $0\leq\lambda\leq1$ provides the balance between bias and variance.
	%Let $Q^{\bm{\pi}}$ denote the true $Q$-function of the current policy.
	%The following theorem indicates that if $\bm{\pi}$ is not far from $\bm{\mu}$, the sequence $Q_k=(\mathcal{T}^{\bm{\pi}}_{\lambda})^kQ$ converges to $Q^{\bm{\pi}}$.
	
	%\begin{theorem}
	%	\label{multi-agent Q(lambda)}
	%	Let $\epsilon=\max_{\bm{\tau}}D_{\rm KL}(\bm{\pi}\|\bm{\mu})[\bm{\tau}]$. Assume that the algorithm satisfies the minimum visit frequency, finite trajectories and bounded step-size assumptions.
	%	If $\lambda < \frac{1-\gamma}{\gamma\sqrt{2\epsilon}}$, repeating $\mathcal{T}^{\bm{\pi}}_{\lambda}$ infinitely from arbitrary $Q_0$ leads to $Q^{\bm{\pi}}$ \rm: $\lim_{k\to\infty}Q_k(\bm{\tau}, \bm{a})=Q^{\bm{\pi}}(\bm{\tau}, \bm{a})$.
	%\end{theorem}
	%Please refer to Appendix~\ref{Proof of Theorem 2} for the detailed proof.
	
	%Since we train joint-policy $\bm{\pi}$ according to the multi-agent version of Equation~\ref{optimization problem}, the implicit constraint $D_{\rm KL}(\bm{\pi}\|\bm{\mu})[\bm{\tau}] \leq \epsilon$ promotes the convergence of $Q$-function in Theorem~\ref{multi-agent Q(lambda)}.
	
	\begin{figure}[t]
		\centering
		\includegraphics[width=5.5in]{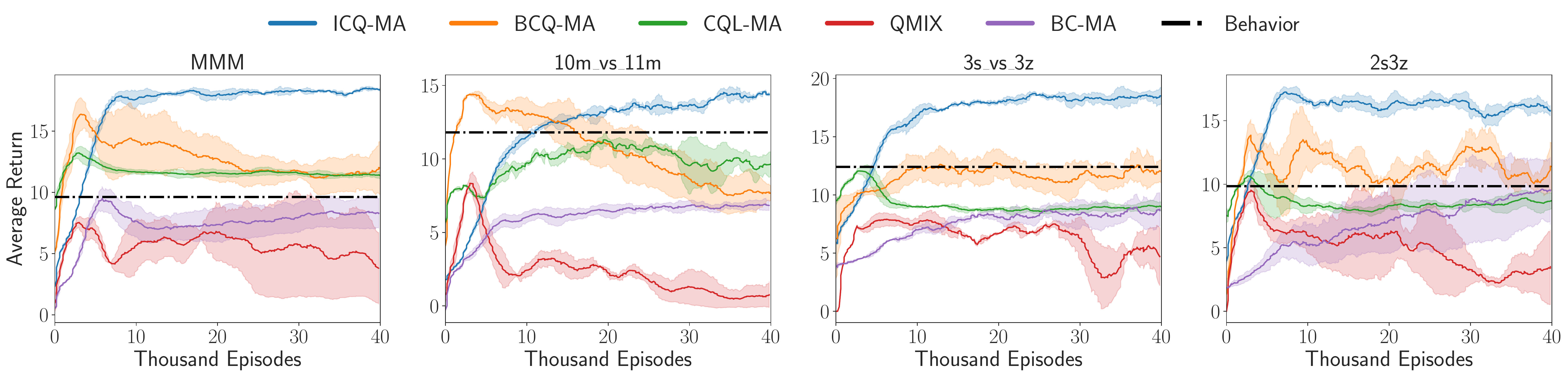}
		\caption{Performance comparison in offline StarCraft II tasks.}
		\label{SMAC result}
	\end{figure}
	
	\begin{table}[t]
		\caption{Performance of ICQ with five offline RL baselines on the single-agent offline tasks with the normalized score metric proposed by D4RL benchmark~\cite{fu2020d4rl}, averaged over three random seeds with standard deviation. Scores roughly range from 0 to 100, where 0 corresponds to a random policy performance and 100 indicates an expert. The results for BC, BCQ, CQL, AWR and BRAC-p are taken from~\cite{fu2020d4rl,kumar2020conservative}.}
		\centering
		\begin{tabular}{cccccccc}
			\toprule
			Dataset type & Environment & ICQ (ours) & BC & BCQ & CQL & AWR & BRAC-p\\
			\midrule
			fixed & antmaze-umaze & $\bm{85.0\pm2.7}$ & 65.0 & 78.9 & 74.0 & 56.0 & 50.0\\
			play & antmaze-medium & $\bm{80.0\pm1.3}$ & 0.0 &  0.0 & 61.2 & 0.0 & 0.0\\
			play & antmaze-large & $\bm{51.0\pm4.8}$ & 0.0 & 6.7 & 15.8 & 0.0 & 0.0\\
			diverse & antmaze-umaze & 65.0$\pm$3.3 & 55.0 & 55.0 & 84.0 & $\bm{70.3}$ & 40.0\\
			diverse & antmaze-medium & $\bm{65.0\pm3.9}$ & 0.0 & 0.0 & 53.7 & 0.0 & 0.0\\
			diverse & antmaze-large & $\bm{44.0\pm4.2}$ & 0.0 & 2.2 & 14.9 & 0.0 & 0.0\\
			\midrule
			expert & adroit-door & $\bm{103.9\pm3.6}$ & 101.2 & 99.0 & - & 102.9 & -0.3\\
			expert & adroit-relocate & $\bm{109.5\pm11.1}$ & 101.3 & 41.6 & - & 91.5 & -0.3\\
			expert & adroit-pen & $\bm{123.8\pm 22.1}$ & 85.1 & 114.9 & - & 111.0 & -3.5\\
			expert & adroit-hammer & $\bm{128.3\pm2.5}$ & 125.6 & 107.2 & - & 39.0 & 0.3\\
			human & adroit-door & 6.4$\pm$2.4 & 0.5 & -0.0 & $\bm{9.1}$ & 0.4 & -0.3\\
			human & adroit-relocate & $\bm{1.5\pm0.7}$ & -0.0 & -0.1 & 0.35 & -0.0 & -0.3\\
			human & adroit-pen & $\bm{91.3\pm10.3}$ & 34.4 & 68.9 & 55.8 & 12.3 & 8.1\\
			human & adroit-hammer & 2.0$\pm$0.9 & 1.5 & 0.5 & $\bm{2.1}$ & 1.2 & 0.3\\
		    \midrule
		    medium & walker2d & 71.8$\pm$10.7 & 66.6 & 53.1 & $\bm{79.2}$ & 17.4 & 77.5\\
		    medium & hopper & 55.6$\pm$5.7 & 49.0 & 54.5 & $\bm{58.0}$ & 35.9 & 32.7\\
		    medium & halfcheetah & 42.5$\pm$1.3 & 36.1 & 40.7 & $\bm{44.4}$ & 37.4 & 43.8\\
		    med-expert & walker2d & $\bm{98.9\pm5.2}$ & 66.8 & 57.5 & 98.7 & 53.8 & 76.9\\
		    med-expert & hopper & 109.0$\pm$13.6 & $\bm{111.9}$ & 110.9 & 111.0 & 27.1 & 1.9\\
		    med-expert & halfcheetah & $\bm{110.3\pm1.1}$ & 35.8 & 64.7 & 104.8 & 52.7 & 44.2\\
 			\bottomrule
		\end{tabular}
		\label{result_D4RL}
	\end{table}

	%\begin{figure}[t]
	%	\centering
	%	\includegraphics[width=5.5in]{fig/D4RL_result.pdf}
	%	\caption{Performance comparison in D4RL.}
	%	\label{D4RL result}
	%\end{figure}
	\section{Related Work}\label{related work}
	As ICQ-MA seems to be the first work addressing the accumulated extrapolation error issue in offline MARL, we briefly review the prior single-agent offline RL works here, which can be divided into three categories: dynamic programming, model-based, and safe policy improvement methods.
	
	\textbf{Dynamic Programming.} Policy constraint methods in dynamic programming~\cite{kalashnikov2018scalable, agarwal2020optimistic, wu2019behavior, todorov2007linearly, jaques2019way} are most closely related to our work.
	They attempt to enforce $\pi$ to be close to $\mu$ under KL-divergence, Wasserstein distance~\cite{vallender1974calculation}, or MMD~\cite{sriperumbudur2012empirical}, and then only use actions sampled from $\pi$ in dynamic programming.
	For example, BCQ~\cite{fujimoto2019off} constrains the mismatch between the state-action visitation of the policy and the state-action pairs contained in the batch by using a state-conditioned generative model to produce only previously seen actions.
	AWR~\cite{peng2019advantage} and ABM~\cite{siegel2020keep} attempt to estimate the value function of the behavior policy via Monte-Carlo or TD($\lambda$).
	Unlike these methods, our algorithm, ICQ, estimates the $Q$-function of the current policy using actions sampled from $\mu$, enabling much more efficient learning.	
	Another series of methods~\cite{touati2020randomized, osband2018randomized, o2018uncertainty} aim to estimate uncertainty to determine the trustworthiness of a $Q$-value prediction.
	However, the high-fidelity requirements for uncertainty estimates limit the performance of algorithms.
	
	\textbf{Model-based and Safe Policy Improvement.} Model-based methods~\cite{jiang2016doubly, thomas2016data, farajtabar2018more, wang2017optimal, kahn2018composable} attempt to learn the model from offline data, with minimal modification to the algorithm.
	Nevertheless, modeling MDPs with very high-dimensional image observations and long horizons is a major open problem, which leads to limited algorithm performance~\cite{levine2020offline}.
	Besides, safe policy improvement methods~\cite{laroche2019safe, sonabend2020expert, berkenkamp2017safe, eysenbach2017leave} require a separately estimated model to $\mu$ to deal with unseen actions.
	However, accurately estimating $\mu$ is especially hard if the data come from multiple sources~\cite{nair2020accelerating}.
	
	\begin{figure}[t]
		\centering
		\subfloat[Performance comparison.]
		{\label{SMAC-Module-Return}\includegraphics[width=2.0in]{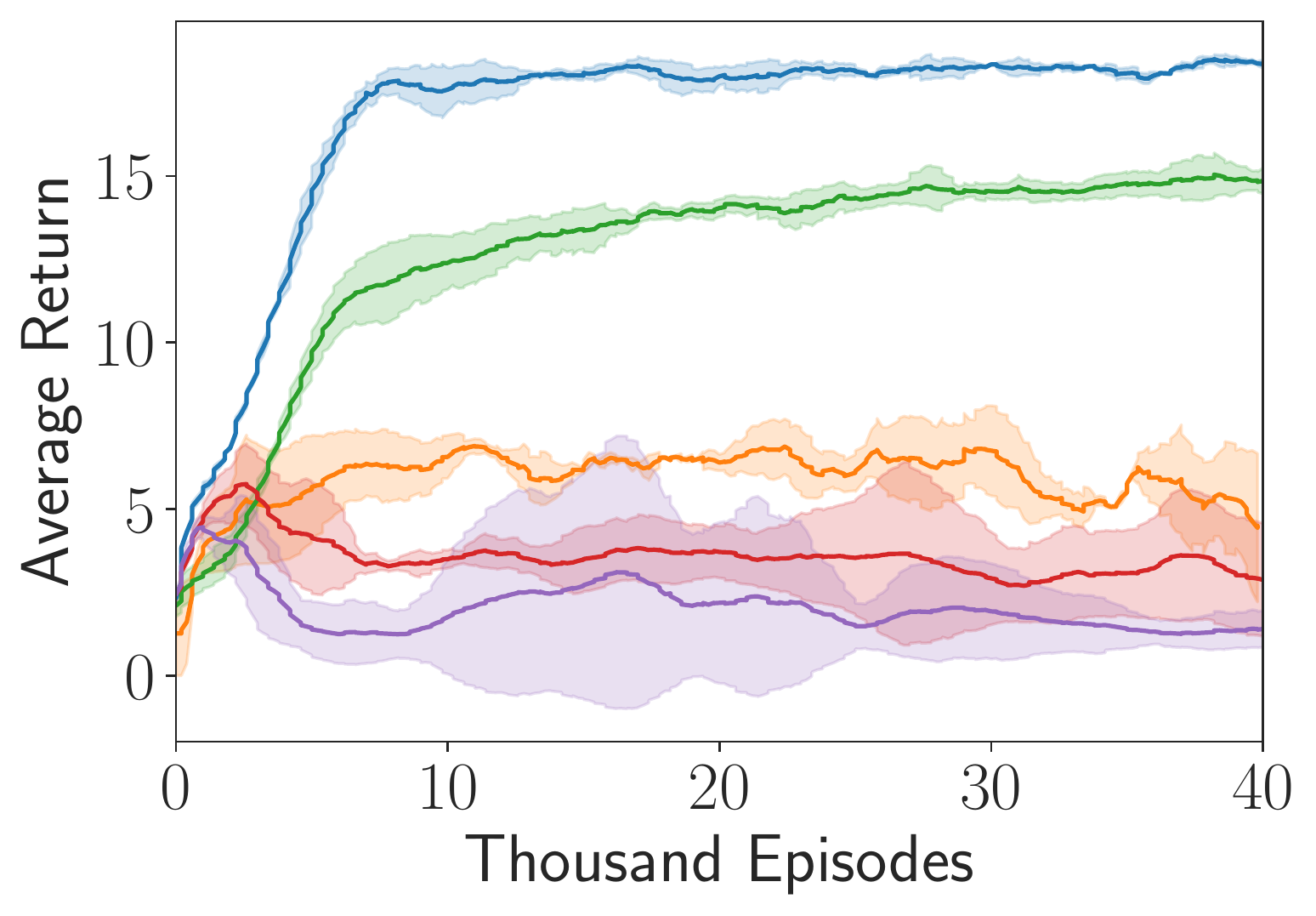}}
		\hspace{1.5mm}
		\subfloat[$Q$-value estimation and True values.]
		{\label{SMAC-Module-Value}\includegraphics[width=3.21in]{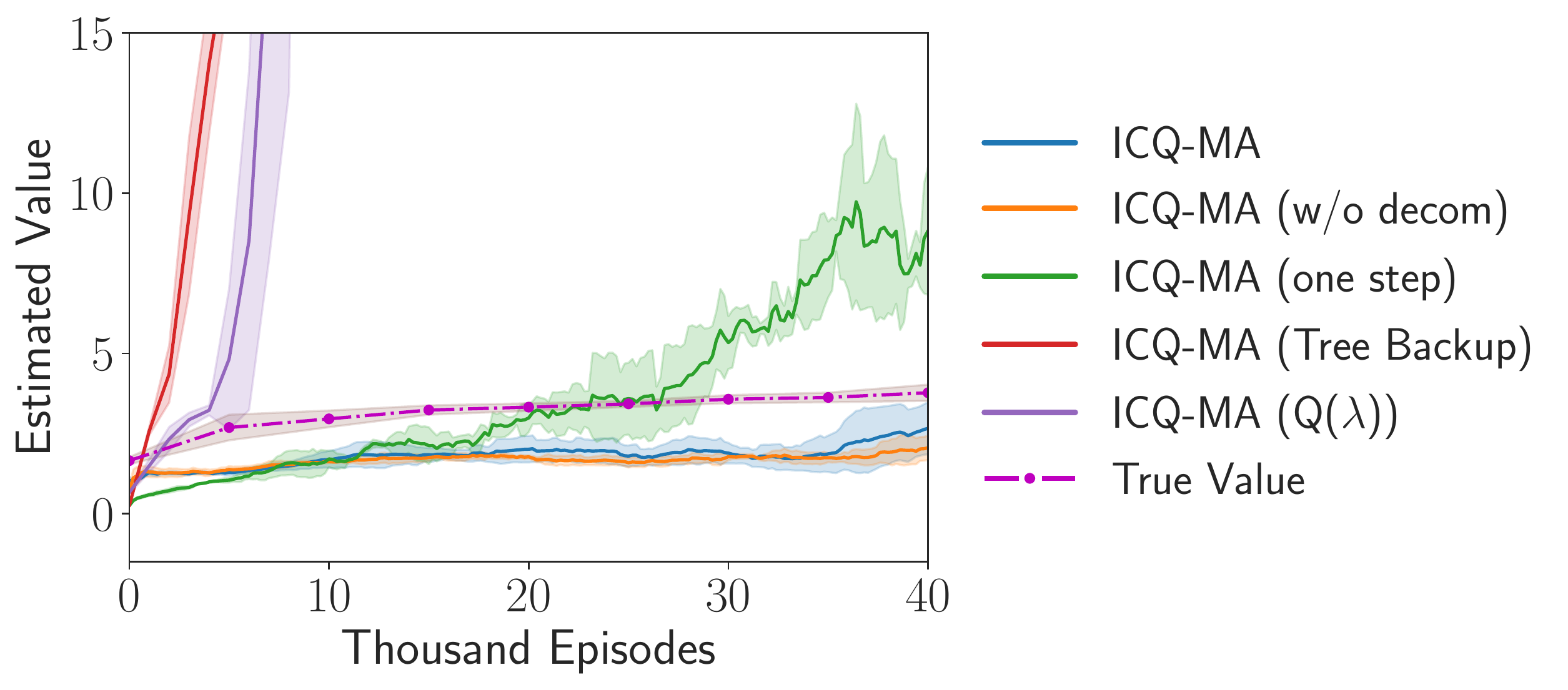}}
		\caption{Module ablation study on MMM map.}
		\label{Ablation Study}
	\end{figure}
	\section{Experiments}\label{experiments}
	In this section, we evaluate ICQ-MA and ICQ on multi-agent~(StarCraft II) and single-agent~(D4RL) offline benchmarks and compare them with state-of-the-art methods.
	Then, we conduct ablation studies on ICQ-MA.
	We aim to better understand each component's effect and further analyze the main driver for the performance improvement.
	
	\subsection{Multi-Agent Offline Tasks on StarCraft II}\label{SMAC}
	
	We first construct the multi-agent offline datasets based on ten maps in StarCraft II~(see
	Table~\ref{Map in SMAC} in Appendix~\ref{Offline MARL Dataset}).
	The datasets are made by collecting DOP~\cite{wang2020off} training data.
	All maps share the same reward function, and each map includes 3000 trajectories.
	We are interested in non-expert data or multi-source data.
	Therefore, we artificially divide behavior policies into three levels based on the average episode return~(see Table~\ref{Behavior policy level} in Appendix~\ref{Offline MARL Dataset}).
	Then, we evenly mix data of three levels.
	
	We compare our method against QMIX~\cite{rashid2018qmix}, multi-agent version of BCQ~(BCQ-MA), CQL~(CQL-MA), and behavior cloning~(BC-MA).
	To maintain consistency, BCQ-MA, CQL-MA, and BC-MA share the same linear value decomposition structure with ICQ-MA.
	Details for baseline implementations are in Appendix~\ref{Baselines Details}.
	Each algorithm runs with five seeds, where the performance is evaluated ten times every 50 episodes.
	Details for hyper-parameters are in Appendix~\ref{hyperparameters}.
	
	We investigate ICQ-MA's performance compared to common baselines in different scenarios.
	Results in Figure~\ref{SMAC result} show that ICQ-MA significantly outperforms all baselines and achieves state-of-the-art performance in all maps.
	QMIX, BCQ-MA, and CQL-MA have poor performances due to the accumulated extrapolation error.
	Interestingly, since BC does not depend on the policy evaluation, it is not subject to extrapolation error.
	Thus BC-MA has a sound performance as StarCraft II is near deterministic.
	We implement BCQ and CQL according to their official code\footnote[1]{BCQ: https://github.com/sfujim/BCQ,	
	\quad CQL: https://github.com/aviralkumar2907/CQL.}.
	
	%\begin{wrapfigure}{R}{0.25\textwidth}
	%    \vspace{-1em}
	%	\includegraphics[width=0.22\textwidth]{fig/D4RL-task.pdf}
	%	\caption{D4RL Task.}
	%    \vspace{-2em}
	%	\label{D4RL-Task}
	%\end{wrapfigure}
	
	\subsection{Single-Agent Offline Tasks on D4RL}\label{D4RL}
	To compare with current offline methods, we evaluate ICQ in the single offline tasks~(e.g., D4RL), including gym domains, Adroit tasks~\cite{rajeswaran2017learning} and AntMaze.
	Specifically, adroit tasks require controlling a 24-DoF robotic hand to imitate human behavior.
	AntMaze requires composing parts of sub-optimal trajectories to form more optimal policies for reaching goals on a MuJoco Ant robot.
	Experimental result in Table~\ref{result_D4RL} shows that ICQ achieves the state-of-the-art performance in many tasks compared with the current offline methods.
	
	\subsection{Ablation Study}\label{Ablation}
	We conduct ablation studies of ICQ-MA in the MMM map of StarCraft II to study the effect of different modules, value estimation, important hyper-parameters, and data quality.
	
	\textbf{Module and Value Estimation Analysis.}
	From Figure~\ref{Ablation Study}, we find that if we adopt other $Q$-value estimation methods in implicit constraint policies~(e.g., $Q(\lambda)$~\cite{munos2016q} or Tree Backup), the corresponding algorithms~(ICQ-MA~($Q(\lambda)$) or ICQ-MA~(Tree Backup)) have poor performances and incorrect estimated values.
	Suppose we train ICQ-MA without decomposed implicit constraint module~(e.g., ICQ-MA ($\rm w/o$ decom)). 
	In that case, the algorithm's performance is poor, although the estimated value is smaller than the true value, confirming the necessity of decomposed policy.
	Besides, the performance of one-step estimation~(ICQ-MA~(one step)) indicates $n$-step estimation is not the critical factor for improving ICQ-MA, while one-step estimation will introduce more bias.

	\textbf{The Parameter $\alpha$.}
	The Lagrangian coefficient $\alpha$ of implicit constraint operator directly affects the intensity of constraint, which is a critical parameter for the performance.
	A smaller $\alpha$ leads to a relaxing constraint and tends to maximize reward.
	If $\alpha \to 0$, ICQ-MA is simplified to $Q$-learning~\cite{watkins1992q} while $\alpha \to \infty$ results in that ICQ-MA is equivalent to behavior cloning.
	Indeed, there is an intermediate value that performs best that can best provide the trade-off as in Appendix~\ref{Appendx:ablation study}.
	
	\textbf{Data Quality.}
	It is also worth studying the performance of ICQ-MA and BC-MA with varying data quality.
	Specifically, we make the datasets from behavior policies of different levels~(e.g., Good, Medium, and Poor).
	As shown in Figure~\ref{Ablation Study_2} in Appendix~\ref{Appendx:ablation study}, ICQ-MA is not sensitive to the data quality, while the performance of BC-MA drops drastically with the data quality deteriorates.
	Results confirm that ICQ-MA is robust to the data quality while BC-MA strongly relies on the data quality.
	
	\textbf{Computational Complexity.}
    With the same training steps in SMAC, BCQ-MA consumes 70\% time of ICQ-MA.
    Although ICQ-MA takes a little long time compared with BCQ-MA, it achieves excellent performance in benchmarks.
    The computing infrastructure for running experiments is a server with an AMD EPYC 7702 64-Core Processor CPU.
	
	\section{Conclusion}\label{conclusion}
	In this work, we demonstrate a critical problem in multi-agent off-policy reinforcement learning with finite data, where it introduces accumulated extrapolation error in the number of agents.
	We empirically show the current offline algorithms are ineffective in the multi-agent offline setting.
	Therefore, we propose the Implicit Constraint Q-learning~(ICQ) method, which effectively alleviates extrapolation error by only trusting the state-action pairs in datasets.
	To the best of our knowledge, the multi-agent version of ICQ is the first multi-agent offline algorithm capable of learning from complex multi-agent datasets.
    % 	We put the discussion of our work's limitation in Appendix~\ref{limitations} as the limited space.
	Due to the importance of offline tasks and multi-agent systems, we sincerely hope our algorithms can be a solid foothold for applying RL to practical applications.

    \begin{ack}
        This work was funded by the National Natural Science Foundation of China (ID:U1813216), National Key Research and Development Project of China under Grant 2017YFC0704100 and Grant 2016YFB0901900, in part by the National Natural Science Foundation of China under Grant 61425027, the 111 International Collaboration Program of China under Grant BP2018006, and BNRist Program (BNR2019TD01009) and the National Innovation Center of High Speed Train R\&D project (CX/KJ-2020-0006).
        
        We sincerely appreciate reviewers, whose valuable comments have benefited our paper significantly!

    \end{ack}
	
	\clearpage
	\bibliographystyle{plain}
	\bibliography{example_paper}

	%%%%%%%%%%%%%%%%%%%%%%%%%%%%%%%%%%%%%%%%%%%%%%%%%%%%%%%%%%%%
	
	\clearpage

	\appendix
	\clearpage
	\setcounter{theorem}{0}
	\setcounter{lemma}{0}
	
	\section{Algorithms}
	\label{algorithm appendix}
	The single-agent version of ICQ is shown in Algorithm~\ref{alg:ICQ single agent}. Its multi-agent version counterpart~(ICQ-MA) is shown in Algorithm~\ref{alg:ICQ multi agent}.
	\begin{algorithm}
		\caption{Implicit Constraint Q-Learning in Single-Agent Tasks.}
		\label{alg:ICQ single agent}
		\KwIn{Offline buffer $\mathcal{B}$, target network update rate $d$.}
		\BlankLine
		Initialize critic network $Q^{\pi}(\cdot;\phi)$ and actor network $\pi(\cdot;\theta)$ with random parameters.\\
		Initialize target networks: $\phi'=\phi$, $\theta'=\theta$.\\
		\For{$t=1$ {\rm \bfseries to} $T$}{
			Sample trajectories from $\mathcal{B}$. \\
			Train policy according to $\mathcal{J}_{\pi}(\theta) = \mathbb{E}_{\tau\sim\mathcal{B}}\left[-\frac{1}{Z(\tau)}\log(\pi(a\mid\tau;\theta))\exp\left(\frac{Q^{\pi}(\tau, a)}{\alpha} \right)\right]$. \\
			Train critic according to $\mathcal{J}_Q(\phi) = \mathbb{E}_{\tau\sim \mathcal{B}}\left[r + \gamma \frac{1}{Z(\tau')}\exp\left(\frac{Q\left(\tau', a'; \phi'\right)}{\alpha}\right) Q\left(\tau', a'; \phi'\right) - Q\left(\tau, a; \phi\right)\right]^2$. \\
			\If{$t$ \rm mod $d=0$}{
				Update target networks: $\phi'=\phi$, $\theta'=\theta$.\\
			}
		}
	\end{algorithm}
	
	\begin{algorithm}
		\caption{Implicit Constraint Q-Learning in Multi-Agent Tasks.}
		\label{alg:ICQ multi agent}
		\KwIn{Offline buffer $\mathcal{B}$, target network update rate $d$.}
		\BlankLine
		Initialize critic networks $Q^{i}(\cdot;\phi_i)$, actor networks $\pi^i(\cdot;\theta_i)$ and Mixer network $M(\cdot; \psi)$ with random parameters.\\
		Initialize target networks: $\phi'=\phi$, $\theta'=\theta$, $\psi' = \psi$.\\
		\For{$t=1$ {\rm \bfseries to} $T$}{
			Sample trajectories from $\mathcal{B}$. \\
			Train individual policy according to $\mathcal{J}_{\bm{\pi}}(\theta) = \sum_i \mathbb{E}_{\tau^i, a^i\sim \mathcal{B}}\left[-\frac{1}{Z^i(\tau^i)}\log(\pi^i(a^i\mid\tau^i;\theta_i))\exp\left(\frac{w^i(\bm{\tau})Q^{i}(\tau^i, a^i)}{\alpha} \right)\right]$. \\
			Train critic according to
			$\mathcal{J}_Q(\phi, \psi)=\mathbb{E}_{ \mathcal{B}}\left[\sum_{t\geq 0}(\gamma\lambda)^t\left[
			r_t + \gamma
			\frac{\exp\left(\frac{1}{\alpha}Q(\bm{\tau}_{t+1},\bm{a}_{t+1};\phi',\psi')\right)}{Z(\bm{\tau}_{t+1};\phi',\psi')}
			Q(\bm{\tau}_{t+1},\bm{a}_{t+1};\phi',\psi')-Q(\bm{\tau}_t, \bm{a}_t;\phi,\psi)\right]
			\right]^2$. \\
			\If{$t$ \rm mod $d=0$}{
				Update target networks: $\phi'=\phi$, $\theta'=\theta$, $\psi'=\psi$.\\
			}
		}
	\end{algorithm}
	
	\clearpage
	\section{Detailed Proof}

\subsection{Proof of Theorem~\ref{theorem_0_appendix}}
    \label{Proof of Theorem 0}	
    \begin{theorem}
		\label{theorem_0_appendix}
		Given a deterministic MDP, the propagation of $\bm{\epsilon_{\rm b}}$ to $\bm{\epsilon_{\rm s}}$ is proportional to $\|P^{\pi}_{\rm s,u}\|_\infty$:
	\begin{equation}
	    \left\| \bm{\epsilon_{\rm s}} \right\|_\infty \leq \frac{\gamma \left\| P^{\pi}_{\rm s, u}\right\|_\infty  }{(1 - \gamma)\left(1 - \gamma \left\|  P^{\pi}_{\rm s, s} \right\|_\infty \right)}\left\| \bm{\epsilon_{\rm b}} \right\|_\infty.
	\end{equation}
	\end{theorem}
	
	\begin{proof}
	Based on the Remark~1 in BCQ~\cite{fujimoto2019off}, the exact form of $\epsilon_{\rm MDP}(\tau, a)$ is:
	\begin{equation}
	\label{MDP error}
	\begin{aligned}
	\epsilon_{\rm MDP}(\tau, a) &= Q_M^{\pi}(\tau, a) - Q^{\pi}_{\mathcal{B}}(\tau, a) \\
	& = \sum_{\tau'}(P_M(\tau'\mid \tau,a) - P_{\mathcal{B}}(\tau'\mid \tau,a)) \left(
	r(\tau,a,\tau') + \gamma\sum_{a'}\pi(a'\mid \tau')Q^{\pi}_{\mathcal{B}}(\tau^\prime, a^\prime)\right)  \\
	& \qquad \qquad + P_M(\tau'\mid \tau,a)\gamma \sum_{a'}\pi(a'\mid \tau')\epsilon_{\rm MDP}(\tau^\prime, a^\prime),
	\end{aligned}
	\end{equation}
	where $P_{\mathcal{B}} = \frac{\mathcal{N}(\tau,a,\tau')}{\sum_{\tilde{\tau}}\mathcal{N}(\tau,a,\tilde{\tau})}$ and $\mathcal{N}$ is the number of times the tuple $(\tau, a, \tau')$ is observed in $\mathcal{B}$.
	If $\sum_{\tilde{\tau}}\mathcal{N}(\tau,a,\tilde{\tau}) = 0$, then $P_{\mathcal{B}}(\tau_{\rm init}\mid \tau, a)=1$.
	Since the considered MDP is deterministic, we have $P_M(\tau'\mid \tau,a) - P_{\mathcal{B}}(\tau'\mid \tau,a) = 0$ for $P^{\pi}_{\rm s,s}$ and $P^{\pi}_{\rm s,u}$.
	For notational simplicity, the error generated by $P_M(\tau'\mid \tau,a) - P_{\mathcal{B}}(\tau'\mid \tau,a)$ in $P^{\pi}_{\rm u,s}$ and $P^{\pi}_{\rm u,u}$ is attributed to $\bm{\epsilon_{\rm b}}$ as they have the same dimension.
	Then, based on the extrapolation error decomposition assumption, we rewrite Equation~\ref{MDP error} in the matrix form:
	\begin{equation}
	\begin{aligned}
	\left[\begin{matrix}
	\bm{\epsilon_{\rm s}} \\
	\bm{\epsilon_{\rm u}}
	\end{matrix}\right] = \gamma  \left[
	\begin{matrix}
	P^{\pi}_{\rm s, s} & P^{\pi}_{\rm s, u} \\
	P^{\pi}_{\rm u, s} & P^{\pi}_{\rm u, u}
	\end{matrix}
	\right] \left[\begin{matrix}
	\bm{\epsilon_{\rm s}} \\
	\bm{\epsilon_{\rm u}}
	\end{matrix}\right] + 
     \left[\begin{matrix}
	\bm{0} \\
	\bm{\epsilon_{\rm b}}
	\end{matrix}\right]
	\end{aligned}.
	\end{equation}
	The result indicates that the error is the solution of a linear program with $[0, \bm{\epsilon_{\rm b}}]^T$ as the reward function. Thus, we solve this linear program and arrive at
	\begin{equation}
	\begin{aligned}
	\left[\begin{matrix}
	\bm{\epsilon_{\rm s}} \\
	\bm{\epsilon_{\rm u}}
	\end{matrix}\right] =& \left( I - \gamma  P^\pi \right)^{-1} 
     \left[\begin{matrix}
	0 \\
	\bm{\epsilon_{\rm b}}
	\end{matrix}\right]
	= \left[
	\begin{matrix}
	I - \gamma P^{\pi}_{\rm s, s} & - \gamma P^{\pi}_{\rm s, u} \\
	- \gamma P^{\pi}_{\rm u, s} & I - \gamma P^{\pi}_{\rm u, u}
	\end{matrix}
	\right]^{-1} 
     \left[\begin{matrix}
	0 \\
	\bm{\epsilon_{\rm b}}
	\end{matrix}\right]
	= \left[
	\begin{matrix}
	A & B \\
	C & D
	\end{matrix}
	\right]^{-1} 
     \left[\begin{matrix}
	0 \\
	\bm{\epsilon_{\rm b}}
	\end{matrix}\right].
	\end{aligned}
	\end{equation}
	With the block matrix inverse formula, we have
    \begin{equation}
    \left[\begin{matrix}
    A & B \\
    C & D
    \end{matrix}\right]^{-1}
    =\left[\begin{matrix}
    A^{-1}+A^{-1} B\left(D-C A^{-1} B\right)^{-1} C A^{-1} & -A^{-1} B\left(D-C A^{-1} B\right)^{-1} \\
    -\left(D-C A^{-1} B\right)^{-1} C A^{-1} & \left(D-C A^{-1} B\right)^{-1}
    \end{matrix}\right]. \label{equ:111}
    \end{equation}
	Since $\left(D-C A^{-1} B\right)^{-1}$ is just the lower right block of $\left( I - \gamma  P^\pi \right)^{-1}$, we have 
	\begin{equation}
	    \left\| \left(D-C A^{-1} B\right)^{-1} \right\|_\infty \leq \left\|\left( I - \gamma  P^\pi \right)^{-1}  \right\|_\infty \leq \frac{1}{1 - \gamma}.
	\end{equation}
	Thus, we obtain
	\begin{equation}
	\begin{aligned}
	    \left\| -A^{-1}B\left(D-C A^{-1} B\right)^{-1} \right\|_\infty \leq& \left\| A^{-1} \right\|_\infty \left\|-B\right\|_\infty \left\|\left(D-C A^{-1} B\right)^{-1} \right\|_\infty  \\
	    \leq& \frac{1}{1 - \gamma}\left\| A^{-1} \right\|_\infty \left\|-B\right\|_\infty \\
	    =& \frac{1}{1 - \gamma}\left\| \left(I - \gamma P^{\pi}_{\rm s, s} \right)^{-1} \right\|_\infty \left\|\gamma P^{\pi}_{\rm s, u}\right\|_\infty\\
	    \leq& \frac{\gamma \left\| P^{\pi}_{\rm s, u}\right\|_\infty  }{(1 - \gamma)\left(1 - \gamma \left\|  P^{\pi}_{\rm s, s} \right\|_\infty \right)}.
	\end{aligned}
	\end{equation}
	Plugging the result into Equation~\ref{equ:111}, we finish our proof at
	\begin{equation}
	    \left\| \bm{\epsilon_{\rm s}} \right\|_\infty \leq \left\|  -A^{-1}B\left(D-C A^{-1} B\right)^{-1} \right\|_\infty \left\| \bm{\epsilon_{\rm b}} \right\|_\infty \leq \frac{\gamma \left\| P^{\pi}_{\rm s, u}\right\|_\infty  }{(1 - \gamma)\left(1 - \gamma \left\|  P^{\pi}_{\rm s, s} \right\|_\infty \right)}\left\| \bm{\epsilon_{\rm b}} \right\|_\infty.
	\end{equation}
	\end{proof}

	\subsection{Proof of Theorem~\ref{ICQ_convergence}}
    The proof of our Theorem 2 is based on the Theorem 3 in~\cite{song2019revisiting}. The main difference is that we consider a behavior policy to regularize the softmax operation. All the actions considered in the analysis are batch-constrained, thus $\mu(a \mid \tau) > 0, \forall \tau, a$ in the proof.
    
	\label{Proof of Theorem 1}
	\begin{lemma}
	\label{Appendix:lemma}
    By assuming $f_{\alpha}^T(Q(\tau,))Q(\tau,)$ as target value of the Implicit Constraint Q-learning operator, we have $\forall Q, \quad \max_{a\sim\mathcal{B}} Q(\tau,a) - f_{\alpha}^T(Q(\tau,))Q(\tau,) \leq (|A_{\tau}|-1)\max\{\frac{1}{(\frac{1}{\alpha}+1)C+1}, \frac{2Q_{\rm max}}{1+C\exp(\frac{1}{\alpha})}\}$, where $Q_{\rm max}=\frac{R_{\rm max}}{1-\gamma}$ represents the maximum $Q$-value in $Q$-iteration with $\mathcal{T}_{\rm ICQ}$.
	\end{lemma}
	\begin{proof}
	The target value operation of Implicit Constraint Q-learning is defined as:
	\begin{equation}
	    \begin{aligned}
	    f_{\alpha}(\tau\mid\mu) = \frac{\left[\mu_1\exp(\frac{1}{\alpha} \tau_1), \mu_2\exp(\frac{1}{\alpha} \tau_2), ..., \mu_{|A_{\tau}|}\exp(\frac{1}{\alpha} \tau_{|A_{\tau}|}) \right]^T}{\sum_{i=1}^{|A_{\tau}|}\mu_i\exp(\frac{1}{\alpha} \tau_i)},
	    \end{aligned}
	\end{equation}
	
	We first sort the sequence $\{Q(\tau,a_i)\}$ such that $Q(\tau,a_{[1]})\ge \cdots\ge Q(\tau,a_{[|A_{\tau}|]})$. Then, $\forall Q$ and $\forall \tau$, we have that the distance between optimal $Q$-value and Implicit Constraint $Q$-value is:
	\begin{equation}
	    \begin{aligned}
	    \label{eq:lemma1_1}
	    &\max_{a\sim\mathcal{B}} Q(\tau,a) - f_{\alpha}^T(Q(\tau,\cdot)\mid\mu(\cdot|\tau))Q(\tau,)\\
	    &=Q(\tau, a_{[1]}) - \frac{\sum_{i=1}^{|A_{\tau}|}\mu(a_{[i]}\mid \tau)\exp\left[\frac{1}{\alpha} Q\left(\tau, a_{[i]}\right)\right]Q\left(\tau,a_{[i]}\right)}{\sum_{i=1}^{|A_{\tau}|}\mu(a_{[i]}\mid \tau)\exp\left[\frac{1}{\alpha} Q\left(\tau, a_{[i]}\right)\right]}\\
	    &= \frac{\sum_{i=1}^{|A_{\tau}|}\mu(a_{[i]}\mid \tau)\exp\left[\frac{1}{\alpha} Q\left(\tau,a_{[i]}\right)\right]\left[Q\left(\tau,a_{[1]}\right)-Q\left(\tau,a_{[i]}\right)\right]}{\sum_{i=1}^{|A_{\tau}|}\mu(a_{[i]}\mid \tau)\exp\left[\frac{1}{\alpha} Q\left(\tau,a_{[i]}\right)\right]}.
	    \end{aligned}
	\end{equation}
	Let $\delta_i(\tau) = Q\left(\tau,a_{[1]}\right)-Q\left(\tau, a_{[i]}\right)$.
	The distance in the Equation~\ref{eq:lemma1_1} can be rewritten as:
	\begin{equation}
	    \begin{aligned}
	    \label{eq:lemma1_3}
	    &\frac{\sum_{i=1}^{|A_{\tau}|}\mu(a_{[i]}\mid \tau)\exp\left[\frac{1}{\alpha} Q\left(\tau, a_{[i]}\right)\right]\left[Q\left(\tau,a_{[1]}\right)-Q\left(\tau,a_{[i]}\right)\right]}{\sum_{i=1}^{|A_{\tau}|}\mu(a_{[i]}\mid \tau)\exp\left[\frac{1}{\alpha} Q\left(\tau,a_{[i]}\right)\right]}\\
	    &=\frac{\sum_{i=1}^{|A_{\tau}|}\mu(a_{[i]}\mid \tau)\exp\left[-\frac{1}{\alpha}\delta_i(\tau)\right]\delta_i(\tau)}{\sum_{i=1}^{|A_{\tau}|}\mu(a_{[i]}\mid \tau)\exp\left[-\frac{1}{\alpha}\delta_i(\tau)\right]}\\
	    &=\frac{\sum_{i=2}^{|A_{\tau}|}\mu(a_{[i]}\mid \tau)\exp\left[-\frac{1}{\alpha}\delta_i(\tau)\right]\delta_i(\tau)}{
	    \mu(a_{[1]}\mid \tau) + \sum_{i=2}^{|A_{\tau}|}\mu(a_{[i]}\mid \tau)\exp\left[-\frac{1}{\alpha}\delta_i(\tau)\right]}
	    \end{aligned}
	\end{equation}
	First note that for any two non-negative sequences $\{x_i\}$ and $\{y_i\}$, 
	\begin{equation}
	    \begin{aligned}
	    \label{eq:lemma1_2}
	    \frac{\sum_ix_i}{1+\sum_iy_i} \leq \sum_i\frac{x_i}{1+y_i}.
	    \end{aligned}
	\end{equation}
    We have the following conclusion by applying the Equation~\ref{eq:lemma1_2} to Equation~\ref{eq:lemma1_3}:
    \begin{equation}
        \begin{aligned}
        \label{Eq:lemma1}
        \frac{\sum_{i=2}^{|A_{\tau}|}\mu(a_{[i]}\mid \tau)\exp\left[-\frac{1}{\alpha}\delta_i(\tau)\right]\delta_i(\tau)}{
	    \mu(a_{[1]}\mid \tau) + \sum_{i=2}^{|A_{\tau}|}\mu(a_{[i]}\mid \tau)\exp\left[-\frac{1}{\alpha}\delta_i(\tau)\right]}
	    &\leq \sum_{i=2}^{|A_{\tau}|} \frac{\mu(a_{[i]}\mid \tau)\exp\left[-\frac{1}{\alpha}\delta_i(\tau)\right]\delta_i(\tau)}{\mu(a_{[1]}\mid \tau) + \mu(a_{[i]}\mid \tau)\exp\left[-\frac{1}{\alpha}\delta_i(\tau)\right]} \\
	    &= \sum_{i=2}^{|A_{\tau}|} \frac{\mu(a_{[i]}\mid \tau)\delta_i(\tau)}{\mu(a_{[i]}\mid \tau) + \mu(a_{[1]}\mid \tau)\exp\left[\frac{1}{\alpha}\delta_i(\tau)\right]} \\
	    &= \sum_{i=2}^{|A_{\tau}|} \frac{\delta_i(\tau)}{1 + \frac{\mu(a_{[1]}\mid \tau)}{\mu(a_{[i]}\mid \tau)}\exp\left[\frac{1}{\alpha}\delta_i(\tau)\right]} \\
	    &\leq \sum_{i=2}^{|A_{\tau}|} \frac{\delta_i(\tau)}{1 + C\exp\left[\frac{1}{\alpha}\delta_i(\tau)\right]},
        \end{aligned}
    \end{equation}
    where $C=\inf_{\tau\in S}\inf_{2\leq i\leq |A_{\tau}|}\frac{\mu(a_{[1]}\mid \tau)}{\mu(a_{[i]}\mid \tau)}$.\\
	If $\delta_i(\tau) > 1$, we have
	\begin{equation}
	    \begin{aligned}
	    \frac{\delta_i(\tau)}{1+C\exp\left[\frac{1}{\alpha} \delta_i(\tau) \right]} \leq \frac{\delta_i(\tau)}{1+C\exp\left(\frac{1}{\alpha}\right)} \leq \frac{2Q_{\rm max}}{1 + C\exp(\frac{1}{\alpha})}.
	    \end{aligned}
	\end{equation}
	else $0 \leq \delta_i(\tau) \leq 1$:
	\begin{equation}
	    \begin{aligned}
	    \frac{\delta_i(\tau)}{1 + C\exp\left[\frac{1}{\alpha}\delta_i(\tau)\right]} = \frac{1}{\frac{1+C}{\delta_i(\tau)} + \frac{1}{\alpha}C + 0.5\frac{1}{\alpha^2}\delta_i(\tau)C + \cdots} \leq \frac{1}{(\frac{1}{\alpha}+1)C + 1}.
	    \end{aligned}
	\end{equation}
	By combining these two cases with Equation~\ref{Eq:lemma1}, we complete the proof.
	\end{proof}
	
	\begin{theorem}
	    Let $\mathcal{T}^k_{\rm ICQ} Q_0$ denote that the operator $\mathcal{T}_{\rm ICQ}$ are iteratively applied over an initial state-action value function $Q_0$ for $k$ times. Then, we have 
	    $\forall (\tau,a)$, $\mathop{\lim\sup}_{k\to \infty}\mathcal{T}_{\rm ICQ}^{k} Q_0(\tau,a) \leq Q^*(\tau,a)$,
	    \begin{equation}
	        \begin{aligned}
	        \mathop{\lim\inf}_{k\to \infty}\mathcal{T}_{\rm ICQ}^{k}Q_0(\tau,a) \ge Q^*(\tau,a) - \frac{\gamma(|A_{\tau}|-1)}{(1-\gamma)}\max\left\{\frac{1}{(\frac{1}{\alpha}+1)C+1}, \frac{2Q_{\rm max}}{1+C\exp(\frac{1}{\alpha})}\right\},
	        \end{aligned}
	    \end{equation}
	    where $|A_{\tau}|$ is the size of seen actions for state $\tau$, $C \triangleq\inf_{\tau\in S}\inf_{2\leq i\leq |A_{\tau}|}\frac{\mu(a_{[1]} \mid \tau)}{\mu(a_{[i]}\mid \tau)}$ and $\mu(a_{[1]} \mid \tau)$ denotes the probability of choosing the expert action according to behavioral policy $\mu$.
	    Moreover, the upper bound of $\mathcal{T}^k_{\rm BCQ} Q_0$ - $\mathcal{T}^k_{\rm ICQ} Q_0$ decays exponentially fast in terms of $\alpha$.
	\end{theorem}
	
	\begin{proof}
	We first conjecture that 
	\begin{equation}
	    \begin{aligned}
	    \label{eq:theorem2_1}
	    \mathcal{T}^k_{\rm BCQ} Q_0(\tau,a) - \mathcal{T}_{\rm ICQ}^k Q_0(\tau,a)\leq \sum_{j=1}^k\gamma^j\zeta,
	    \end{aligned}
	\end{equation}
	where $\zeta=\sup_Q\max_\tau\left[\max_{a\sim\mathcal{B}} Q(\tau,a) - f_{\alpha}^T\left(Q(\tau,)\right)Q(\tau,)\right]$ denotes the supremum of the difference between the BCQ and ICQ operators, over all $Q$-functions that occur during $Q$-iteration, and state $\tau$. Equation~\ref{eq:theorem2_1} is proven using induction as follows:
	\begin{itemize}
	    \item When $i=1$, we start from the definitions for $\mathcal{T}_{\rm BCQ}$ and $\mathcal{T}_{\rm ICQ}$, and proceed as 
	    \begin{equation}
	        \begin{aligned}
	        \mathcal{T}_{\rm BCQ}Q_0(\tau,a) - \mathcal{T}_{\rm ICQ}Q_0(\tau,a)
	        &= \gamma\sum_{\tau'}P(\tau'\mid \tau,a)\left[\max_{a'\sim\mathcal{B}}Q_0(\tau',a') - f_{\alpha}^T\left(Q_0(\tau',)\right)Q_0(\tau',) \right]\\
	        &\leq \gamma\sum_{\tau'}P(\tau'\mid \tau, a)\zeta = \gamma\zeta.
	        \end{aligned}
	    \end{equation}
	    \item Suppose the conjecture holds when $i=l$, i.e., $\mathcal{T}^l_{\rm BCQ} Q_0(\tau, a) - \mathcal{T}^l_{\rm ICQ} Q_0(\tau, a)\leq\sum_{j=1}^{l}\gamma^j\zeta$, then
	    \begin{equation}
	        \begin{aligned}	   \mathcal{T}_{\rm BCQ}^{l+1}Q_0(\tau,a) - \mathcal{T}^{l+1}_{\rm ICQ} Q_0(\tau,a) 
	       &=\mathcal{T}_{\rm BCQ}\mathcal{T}_{\rm BCQ}^{l}Q_0(\tau,a) - \mathcal{T}^{l+1}_{\rm ICQ}Q_0(\tau,a) \\
	        &\leq \mathcal{T}_{\rm BCQ}\left[\mathcal{T}^{l}_{\rm ICQ}Q_0(\tau,a) + \sum_{j=1}^l\gamma^j\zeta\right] - \mathcal{T}^{l+1}_{\rm ICQ}Q_0(\tau,a)\\
	        &= \sum_{j=1}^l\gamma^{j+1}\zeta + \left(\mathcal{T}_{\rm BCQ} - \mathcal{T}_{\rm ICQ}\right)\mathcal{T}^{l}_{\rm ICQ}Q_0(\tau, a)\\
	        &\leq \sum_{j=1}^l\gamma^{j+1}\zeta + \gamma\zeta = \sum_{j=1}^{l+1}\gamma^j\zeta.
	        \end{aligned}
	    \end{equation}
	\end{itemize}
	By using the fact that $\lim_{k\to\infty}\mathcal{T}_{\rm BCQ}^kQ_0(\tau,a)$ and applying Lemma~\ref{Appendix:lemma} to bound $\zeta$, we have $\forall (\tau,a)$, $\lim\sup_{k\to \infty}\mathcal{T}_{\rm ICQ}^{k}Q_0(\tau,a) \leq Q^*(\tau,a)$ and $\lim\inf_{k\to \infty}\mathcal{T}_{\rm ICQ}^{k}Q_0(\tau,a) \ge Q^*(\tau,a) - \frac{\gamma(|A_{\tau}|-1)}{(1-\gamma)}\max\{\frac{1}{(\frac{1}{\alpha}+1)C+1}, \frac{2Q_{\rm max}}{1+C\exp(\frac{1}{\alpha})}\}$.
    Based on the Equation~\ref{Eq:lemma1}, we can bound Equation~\ref{eq:theorem2_1} as:
    \begin{equation}
        \begin{aligned}
        \label{eq:theorem2_2}
        \mathcal{T}_{\rm BCQ}^kQ_0(\tau,a) - \mathcal{T}_{\rm ICQ}^kQ_0(\tau,a) \leq \frac{\gamma(1-\gamma^k)}{1-\gamma}\sum_{i=2}^{|A_{\tau}|} \frac{\delta_i(\tau)}{1 + C\exp\left[\frac{1}{\alpha}\delta_i(\tau)\right]}.
        \end{aligned}
    \end{equation}
	From the definition of $\delta_i(\tau)$, we have $\delta_{|A_{\tau}|}(\tau) \ge \delta_{|A_{\tau}|-1}(\tau) \ge \cdots \ge \delta_2(\tau) \ge 0$. 
	Furthermore, there must exist an index $i^* \leq |A_{\tau}|$ such that $\delta_i > 0, \forall i^*\leq i \leq |A_{\tau}|$.
	Therefore, we can proceed from Equation~\ref{eq:theorem2_2} as
	\begin{equation}
	    \begin{aligned}
	    \frac{\gamma(1-\gamma^k)}{1-\gamma}&\sum_{i=2}^{|A_{\tau}|} \frac{\delta_i(\tau)}{1 + C\exp\left[\frac{1}{\alpha}\delta_i(\tau)\right]}
	    =\frac{\gamma(1-\gamma^k)}{1-\gamma}\sum_{i=i^*}^{|A_{\tau}|} \frac{\delta_i(\tau)}{1 + C\exp\left[\frac{1}{\alpha}\delta_i(\tau)\right]}\\
	    &\leq \frac{\gamma(1-\gamma^k)}{1-\gamma}\sum_{i=i^*}^{|A_{\tau}|} \frac{\delta_i(\tau)}{C\exp\left[\frac{1}{\alpha}\delta_i(\tau)\right]}
	    \leq \frac{\gamma(1-\gamma^k)}{1-\gamma}\sum_{i=i^*}^{|A_{\tau}|} \frac{\delta_i(\tau)}{C\exp\left[\frac{1}{\alpha}\delta_{i^*}(\tau)\right]}\\
	    &= \frac{\gamma(1-\gamma^k)}{1-\gamma}\exp\left[-\frac{1}{\alpha}\delta_{i^*}(\tau)\right]\sum_{i=i^*}^{|A_{\tau}|} \frac{\delta_i(\tau)}{C},
	    \end{aligned}
	\end{equation}
	which implies an exponential convergence rate in terms of $\alpha$.
	\end{proof}
	
	\subsection{Proof of Remark~\ref{Toy Example}}
	We analyze the MMDP experimental result in Section~\ref{Toy Example} from the perspective of the concentrability coefficient $C(\Pi)$, which illustrates the degree to which states and actions are out of distribution.
	In the MMDP case, we theoretically prove $C(\Pi^i)$ satisfies: $C(\Pi^1) < C(\Pi^2) <\dots < C(\Pi^n)$, where $\Pi^i$ denotes the set of joint policies including $i$ agents.
	As illustrated in the above conclusion, the increase in the number of agents makes the distribution shift issue more severe in the MMDP case.
	\begin{remark}
		\label{theorem_1_appendix}
		Let $\varrho(s)$ denote the marginal distribution over $S$, $\rho_0$ indicate the initial state distribution, and $\Pi^i$ represent the set of joint policies including $i$ agents.
		Assume there exist coefficients $c(k)$ satisfying $\rho_0 P^{\bm{\pi}_1} P^{\bm{\pi}_2} ... P^{\bm{\pi}_k}(s) \leq c(k) \varrho(s)$.
		We define the concentrability coefficient $C(\Pi) \triangleq (1-\gamma)^2\sum_{k=1}^{\infty}k\gamma^{k-1}c(k)$, which illustrates the degree to which states and actions are out of distribution.
		Due to the limited datasets, the number of seen state-action pairs $m$ is fixed.
		Then, $C(\Pi^i)$ is monotonically increasing with the number of agents
		\begin{equation}
		C(\Pi^1) < C(\Pi^2) <\dots < C(\Pi^n)
		\end{equation}
	\end{remark}
	\begin{proof}
	We first note that $c(k) \geq \frac{\rho_0 P^{\bm{\pi}_1} P^{\bm{\pi}_2} ... P^{\bm{\pi}_k}(s)}{\varrho(s)}$ and $c(k)$ determines the value of $C(\Pi^i)$.
	To compare $C(\Pi^i)$, we just need to compare $c(k)$ at iteration $k$.
	For clarity of analysis, we assume each state-action pair is visited only once, and individual policy is random $\pi^i(A^{(i)}|s)=\frac{1}{2}$.
	In the MMDP case, the transition matrix $P^{\bm{\pi}}$ is stable for the number of agents:
	\begin{equation}
	\begin{aligned}
	P^{\pi^1_k} = P^{\bm{\pi}_k} = P^{\pi^1_k \pi^2_k \dots \pi^n_k} = \left[\begin{matrix}
	1 & 0\\
	\frac{1}{2} & \frac{1}{2}
	\end{matrix}\right]
	\end{aligned}.
	\end{equation}
	For this reason, $\rho_0 P^{\bm{\pi}_1} P^{\bm{\pi}_2} ... P^{\bm{\pi}_k}(s)$ does not change with the number of agents. 
	As $\varrho(s) = \sum_{a}\varrho(s,a) = \sum_a \frac{\sum_{s,\bm{a}\in \mathcal{D}} \mathbf{1}[s=s, \bm{a}=\bm{a}]}{\sum_{s',\bm{a}'\in \mathcal{D}} \mathbf{1}[s=s', \bm{a}=\bm{a}']}$, we can calculate $\varrho(s)$ by counting state-action pairs in $\mathcal{D}$ as follows
	\begin{equation}
	\begin{aligned}
	\varrho(s) = \frac{m}{2^{n+1}}
	\end{aligned}.
	\end{equation}
	The gradient of $\varrho(s)$ is:
	\begin{equation}
	\varrho(s)' = \left(\frac{m}{2^{n+1}} \right)'=\frac{-m\cdot 2^{n+1}\ln 2}{(2^{n+1})^2} < 0.
	\end{equation}
	Therefore, $c(k)$ is monotonically increasing with the number of agents and $C(\Pi^1) < C(\Pi^2) <\dots < C(\Pi^n)$.
	\end{proof}
	
	\subsection{Proof of Remark~\ref{theorem_1.5_appendix}}
	\label{Proof of Theorem 1.5}
	\begin{remark}
		\label{theorem_1.5_appendix}
		For the optimization problem 
		\begin{equation}
		\label{optimization problem in appendix}
		\begin{aligned}
	    \pi_{k+1} = \mathop{\arg\max}_{\pi}\mathbb{E}_{a\sim\pi(\cdot\mid\tau)}[Q^{\pi_k}(\tau, a)] \qquad{\rm s.t. \quad}  D_{\rm KL}(\pi \|\mu)[\tau] \leq \epsilon, \quad \sum_a\pi(a | \tau) = 1,
		\end{aligned}
		\end{equation}
		the optimal policy is $\pi^*_{k+1}(a\mid\tau) = \frac{\mu(a\mid\tau)\exp\left(\frac{1}{\alpha}Q^{\pi_k}(\tau, a)\right)}{\sum_{\tilde{a}} \mu(\tilde{a}\mid\tau)\exp\left(\frac{1}{\alpha}Q^{\pi_k}(\tau, \tilde{a})\right)}$.
	\end{remark}
	\begin{proof}
	First, note the objective is a linear function of the decision variables $\pi$. All constraints are convex functions. Thus Equation~\ref{optimization problem in appendix} is a convex optimization problem.
	The Lagrangian equation is
	\begin{equation}
	\begin{aligned}
	\label{Lagrangian}
	\mathcal{L}(\pi, \alpha) = \mathbb{E}_{a\sim\pi}[Q^{\pi_k}(\tau, a)] + \alpha\left(\epsilon - D_{\rm KL}\left(\pi \|\mu\right)[\tau]\right) + \lambda\left(1- \sum_a\pi(a\mid\tau)\right),
	\end{aligned}
	\end{equation}
	where $\alpha > 0$ denotes the Lagrangian coefficient.
	Differentiate $\pi$ to get the following formula
	\begin{equation}
	\begin{aligned}
	\frac{\partial \mathcal{L}}{\partial \pi}=Q^{\pi_k}(\tau, a) - \alpha \left(1+\log\left(\frac{\pi(a\mid\tau)}{\mu(a\mid\tau)}\right)\right) - \lambda.
	\end{aligned}
	\end{equation}
	Setting $\frac{\partial \mathcal{L}}{\partial \pi}$ to zero, then:
	\begin{equation}
	    \begin{aligned}
	    &Q^{\pi_k}(\tau, a) - \alpha \left(1+\log\left(\frac{\pi(a\mid\tau)}{\mu(a\mid\tau)}\right)\right) - \lambda = 0 \\
	    &Q^{\pi_k}(\tau, a) = \alpha \left(1+\log\left(\frac{\pi(a\mid\tau)}{\mu(a\mid\tau)}\right)\right) + \lambda \\
	    &\frac{Q^{\pi_k}(\tau, a)}{\alpha} - 1 - \frac{\lambda}{\alpha} = \log\left(\frac{\pi(a\mid\tau)}{\mu(a\mid\tau)}\right)\\
	    &\frac{\pi(a \mid \tau)}{\mu(a\mid\tau)} = \exp\left(\frac{Q^{\pi_k}(\tau, a)}{\alpha} - 1 - \frac{\lambda}{\alpha}\right)\\
	    &\pi(a\mid\tau) = \mu(a\mid\tau)\exp\left(\frac{Q^{\pi_k}(\tau, a)}{\alpha} - 1 - \frac{\lambda}{\alpha}\right)
	    \end{aligned}
	\end{equation}
	Due to the second constraint in Equation~\ref{optimization problem in appendix}, the policy is a probability distribution.
	Therefore, we adopt $Z$ to normalize the result by moving the constant $\mu(a\mid\tau)\exp(-1-\frac{\lambda}{\alpha})$ to $Z$:
	\begin{equation}
	\begin{aligned}
	\pi^*_{k+1}(a\mid\tau) = \frac{1}{Z(\tau)}\mu(a\mid\tau)\exp\left(\frac{Q^{\pi_k}(\tau, a)}{\alpha}\right),
	\end{aligned}
	\end{equation}
	where $Z(\tau) = \sum_{\tilde{a}}\mu(\tilde{a}\mid\tau)\exp\left(\frac{1}{\alpha}Q^{\pi_k}(\tau, \tilde{a})\right)$ is the normalizing partition function.
	\end{proof}

	\subsection{Proof of Theorem~\ref{theorem_3_appendix}}
	\label{Proof of Theorem 3}
	\begin{theorem}\label{theorem_3_appendix}
		Assuming the joint action-value function is linearly decomposed, we can decompose the multi-agent joint-policy under implicit constraint as follows
		\begin{equation}
		\begin{aligned}
		\bm{\pi} = \mathop{\arg\max}_{\pi^1,\dots,\pi^n} \sum_i \mathbb{E}_{\tau^i, a^i\sim \mathcal{B}}\left[\frac{1}{Z^i(\tau^i)}
		\log (\pi^i(a^i\mid \tau^i)) \exp\left(\frac{w^i(\bm{\tau})Q^i(\tau^i, a^i)}{\alpha}\right)\right]
		\end{aligned},
		\end{equation}
		where $Z^i(\tau^i) = \sum_{\tilde{a}^i}\mu^i(\tilde{a}^i \mid \tau^i) \exp\left(\frac{1}{\alpha}w^i(\bm{\tau})Q^i(\tau^i, \tilde{a}^i)\right)$ is the normalizing partition function.
	\end{theorem}
	\begin{proof}
	Let $\mathcal{J}_{\bm{\pi}}$ denote the joint-policy loss.
	According to the assumption, $\mathcal{J}_{\bm{\pi}}$ is written:
	
	\begin{equation}
	\begin{aligned}
	\mathcal{J}_{\bm{\pi}} & = \mathbb{E}_{\bm{\tau}, \bm{a}\sim \mathcal{B}}\left[-\frac{1}{Z(\bm{\tau})} \log(\bm{\pi}(\bm{a}|\bm{\tau}))\exp\left(\frac{1}{\alpha}Q^{\bm{\pi}}(\bm{\tau}, \bm{a}) \right) \right] \\
	& = \mathbb{E}_{\bm{\tau},a^1,\dots,a^n\sim \mathcal{B}}\left[-\frac{1}{Z(\bm{\tau})}\left(\sum_i \log(\pi^i(a^i\mid\tau^i))\right)
	\exp\left(\frac{1}{\alpha}\left(\sum_i w^i(\bm{\tau})Q^i(\tau^i,a^i) + b(\bm{\tau})
	\right)\right)\right].
	\end{aligned}
	\end{equation}
	The loss function $\mathcal{J}_{\bm{\pi}}$ is equivalent to the following form by relocating the sum operator:
	\begin{equation}
	\begin{aligned}
	\mathcal{J}_{\bm{\pi}} & = \sum_i \mathbb{E}_{\bm{\tau},a^1,\dots,a^n\sim \mathcal{B}}\left[-\frac{1}{Z(\bm{\tau})}\log (\pi^i(a^i\mid\tau^i))
	\exp\left(\frac{\sum_i w^i(\bm{\tau})Q^i(\tau^i,a^i) + b(\bm{\tau})}{\alpha}\right)\right] \\
	& = \sum_i \mathbb{E}_{\bm{\tau},a^1,\dots,a^n\sim \mathcal{B}}
	[-\frac{1}{Z(\bm{\tau})}\log(\pi^i(a^i\mid\tau^i))
	\exp\left(\frac{w^i(\bm{\tau})Q^i(\tau^i,a^i)}{\alpha}\right) \\
	& \qquad \qquad \qquad \qquad \qquad \qquad \qquad \qquad \qquad
	\exp\left(\frac{\sum_{j\ne i} w^j(\bm{\tau})Q^j(\tau^j,a^j) + b(\bm{\tau})}{\alpha}\right)] \\
	& = \sum_i \mathbb{E}_{\bm{\tau},a^i\sim \mathcal{B}}
	\mathbb{E}_{a^{j\ne i}\sim \mathcal{B}}
	[-\frac{1}{Z(\bm{\tau})}\log(\pi^i(a^i\mid\tau^i))
	\exp\left(\frac{w^i(\bm{\tau})Q^i(\tau^i,a^i)}{\alpha}\right) \\
	& \qquad \qquad \qquad \qquad \qquad \qquad \qquad \qquad \qquad
	\exp\left(\frac{\sum_{j\ne i} w^j(\bm{\tau})Q^j(\tau^j,a^j) + b(\bm{\tau})}{\alpha}\right)] \\
	& = \sum_i \mathbb{E}_{\bm{\tau},a^i\sim \mathcal{B}}\left[-\frac{1}{Z^i(\tau^i)}\log( \pi^i(a^i\mid\tau^i))
	\exp\left(\frac{w^i(\bm{\tau})Q^i(\tau^i,a^i)}{\alpha}\right)\right],
	\end{aligned}
	\end{equation}
	\begin{equation}
	\begin{aligned}
	Z^i(\tau^i) &= \frac{\sum_{\tilde{a}^i}\sum_{\tilde{a}^{j\ne i}}\bm{\mu}(\bar{\bm{a}}\mid\bm{\tau})
		\exp\left(\frac{1}{\alpha}w^i(\bm{\tau})Q^i(\tau^i,\tilde{a}^i)\right)\exp\left(\frac{1}{\alpha}(\sum_{j\ne i} w^j(\bm{\tau})Q^j(\tau^j,\tilde{a}^j) + b(\bm{\tau}))\right)}
	{\mathbb{E}_{\tilde{a}^{j\ne i}\sim\mathcal{B}}\left[
		\exp \left(\frac{1}{\alpha}\left(\sum_{j\ne i} w^j(\bm{\tau})Q^j(\tau^j,\tilde{a}^j) + b(\bm{\tau})\right)	\right)\right]} \\
	&= \frac{\sum_{\tilde{a}^i}\sum_{\tilde{a}^{j\ne i}}
		\mu^i(\tilde{a}^i\mid \tau^i)\mu^{j\ne i}(\tilde{a}^j\mid \tau^j)
		\exp\left(\frac{1}{\alpha}w^i(\bm{\tau})Q^i(\tau^i,\tilde{a}^i)\right)}
	{\sum_{\tilde{a}^{j\ne i}}
		\mu^{j\ne i}(\tilde{a}^j\mid \tau^j)	\exp \left(\frac{1}{\alpha}\left(\sum_{j\ne i} w^j(\bm{\tau})Q^j(\tau^j,\tilde{a}^j) + b(\bm{\tau})\right)	\right)
	} \cdot \\
	& \qquad \qquad \qquad \qquad \qquad \qquad \qquad \qquad \qquad
	\exp\left(\frac{1}{\alpha}\left(\sum_{j\ne i} w^j(\bm{\tau})Q^j(\tau^j,\tilde{a}^j) + b(\bm{\tau})\right)\right) \\
	& = \sum_{\tilde{a}^i}\mu^i(\tilde{a}^i \mid \tau^i) \exp\left(\frac{1}{\alpha}w^i(\bm{\tau})Q^i(\tau^i, \tilde{a}^i)\right).
	\end{aligned}
	\end{equation}
	\end{proof}
	
	%\clearpage
	\section{Additional Results}\label{additional results}
	
	%\subsection{Additional Results in StarCraft II}\label{Additional Results in StarCraft II}
	%\begin{figure}[h]
	%	\centering
	%	\includegraphics[width=5.0in]{fig/SMAC_Appendix.pdf}
	%	\caption{Performance comparison in additional StarCraft II environments.}
	%	\label{Additional Results SMAC}
	%\end{figure}
	
	%\subsection{Additional Value Estimation Experiments in StarCraft II}\label{Additional Q-estimates in StarCraft II}
	%We compare the $Q$-estimates between ICA-MA and BCQ-MA, which is shown in Figure~\ref{Additional Q-Results SMAC}.
	%The experimental results demonstrate that the extrapolation error of ICQ-MA is reduced to almost zero and insensitive to the number of agents.
	%However, the $Q$-estimates of BCQ-MA gradually diverge.
	
	%\begin{figure}[h]
	%	\centering
	%	\includegraphics[width=5.0in]{fig/Q-estimates_Appendix.pdf}
	%	\caption{The comparison of estimated $Q$-value in StarCraft II environments.}
	%	\label{Additional Q-Results SMAC}
	%\end{figure}
	
	%\subsection{Additional Results in D4RL}\label{Additional Results in D4RL}
	%\begin{figure}[h]
	%	\centering
	%	\includegraphics[width=5.0in]{fig/d4rl-appendix.pdf}
	%	\caption{Performance comparison in additional D4RL environments.}
	%	\label{Additional Results D4RL}
	%\end{figure}
	
	\subsection{Additional Ablation Results in StarCraft II}\label{Additional Ablation Results}
	\begin{figure}[h]
		\centering
		\includegraphics[width=5.0in]{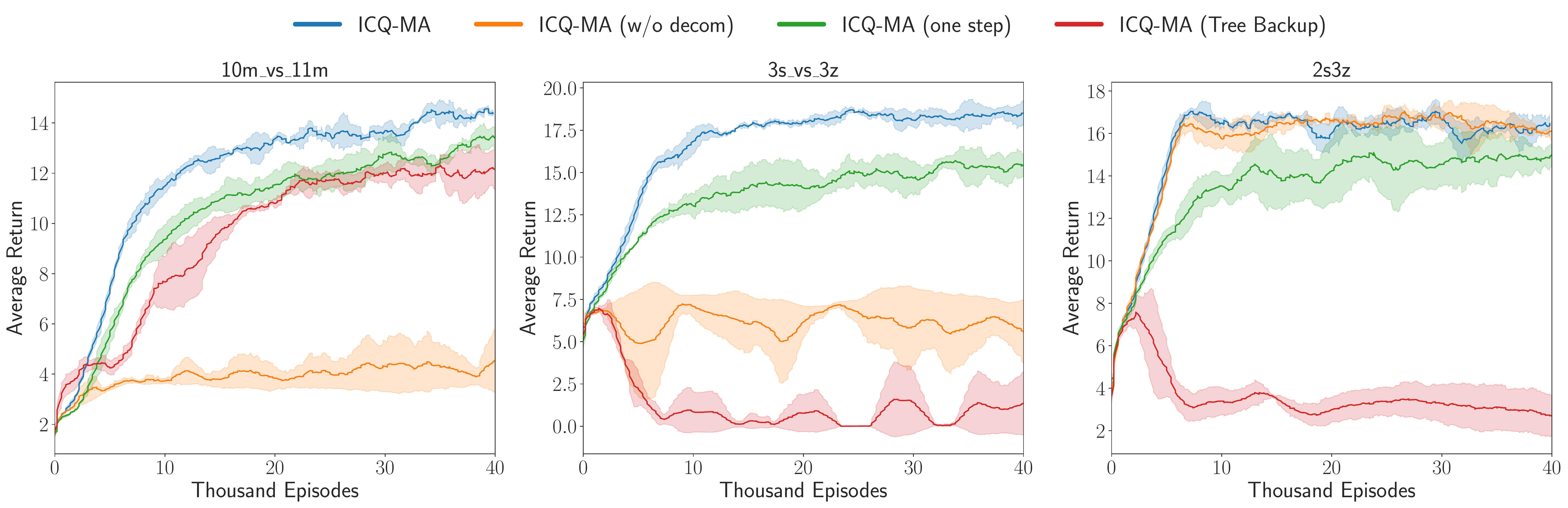}
		\caption{Module ablation study in additional StarCraft II environments.}
		\label{Additional Results ablation}
	\end{figure}	

	\subsection{Additional Results in MMDP}\label{MMDP addtional results}
\begin{figure}[h]
		\centering
		\includegraphics[width=4.8in]{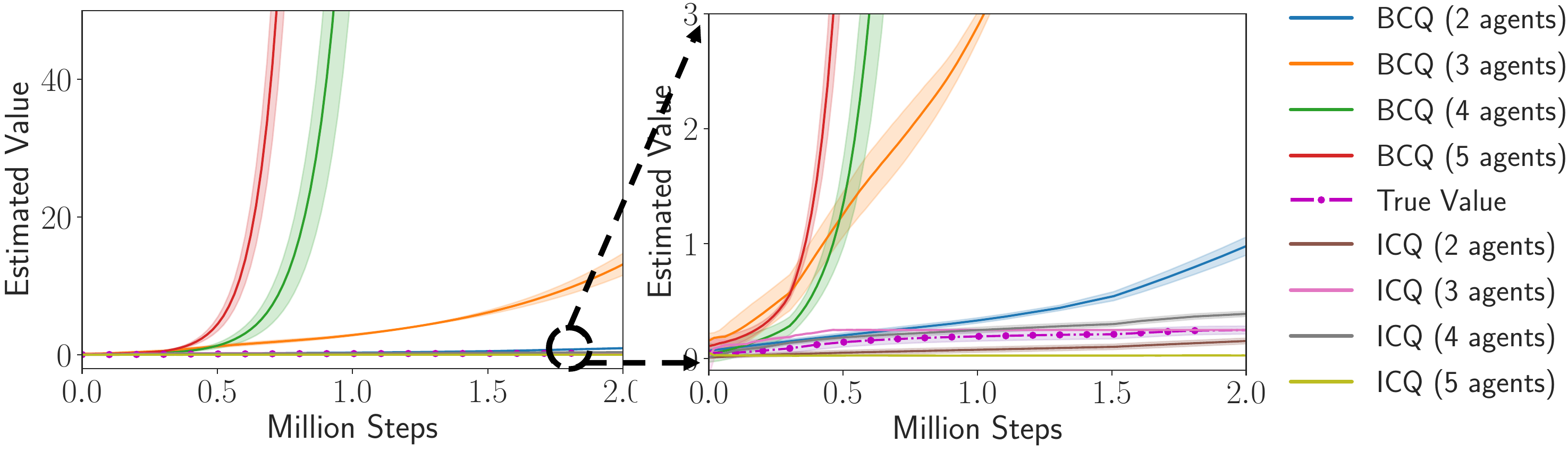}
		\caption{Additional results in MMDP.}
		\label{Ablation results in MMDP}
	\end{figure}

	Due to the space limits, we put the complete results in MMDP in Figure~\ref{Ablation results in MMDP}.
	BCQ gradually diverges as the number of agents increases, while ICQ has accurate estimates.

    \subsection{Additional Results in D4RL}\label{additional results in D4RL}
    \begin{figure}[h]
		\centering
		\subfloat[Adroit-expert.]
		{\includegraphics[width=2.3in]{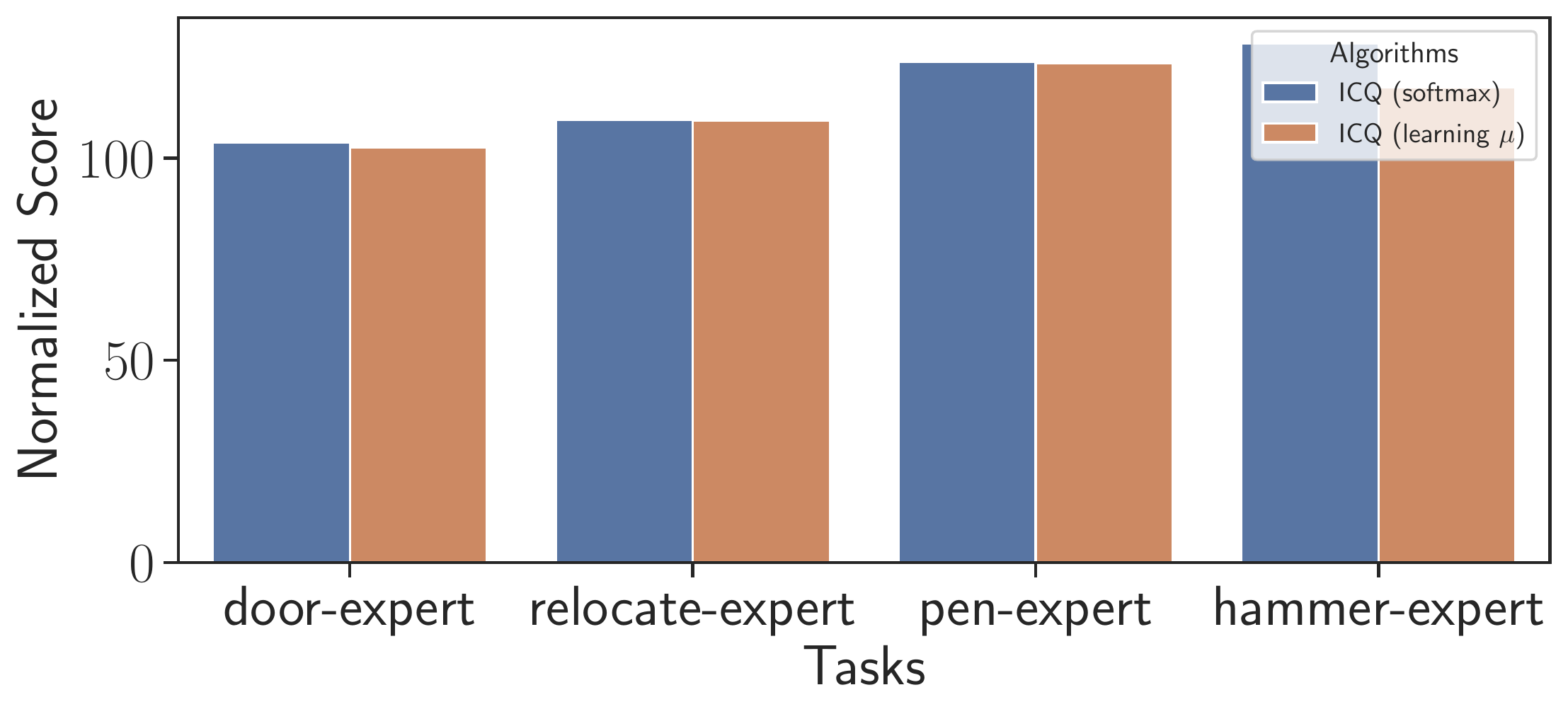}}
		\hspace{1.5mm}
		\subfloat[Adroit-human.]
		{\includegraphics[width=2.3in]{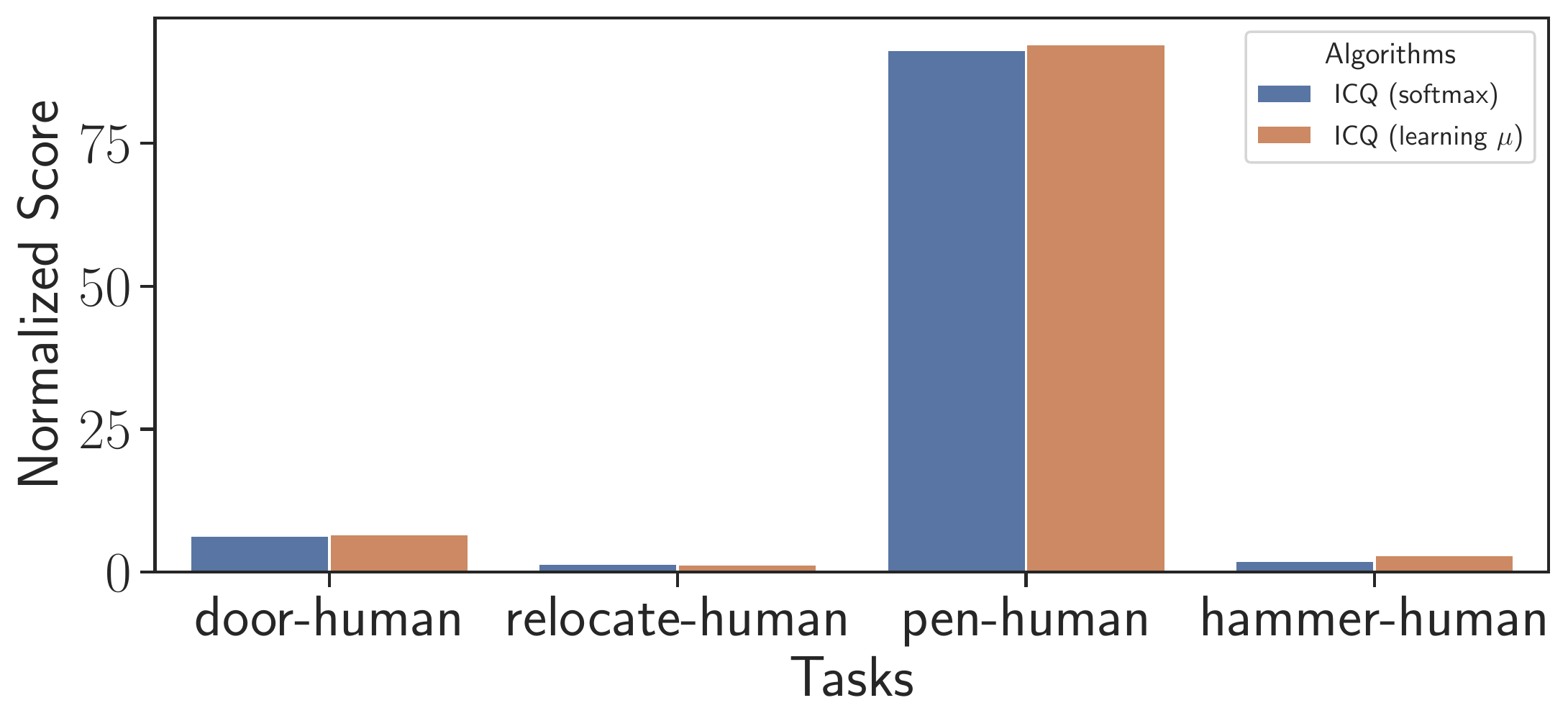}}\\
		\subfloat[Mujoco-medium-expert.]
		{\includegraphics[width=2.3in]{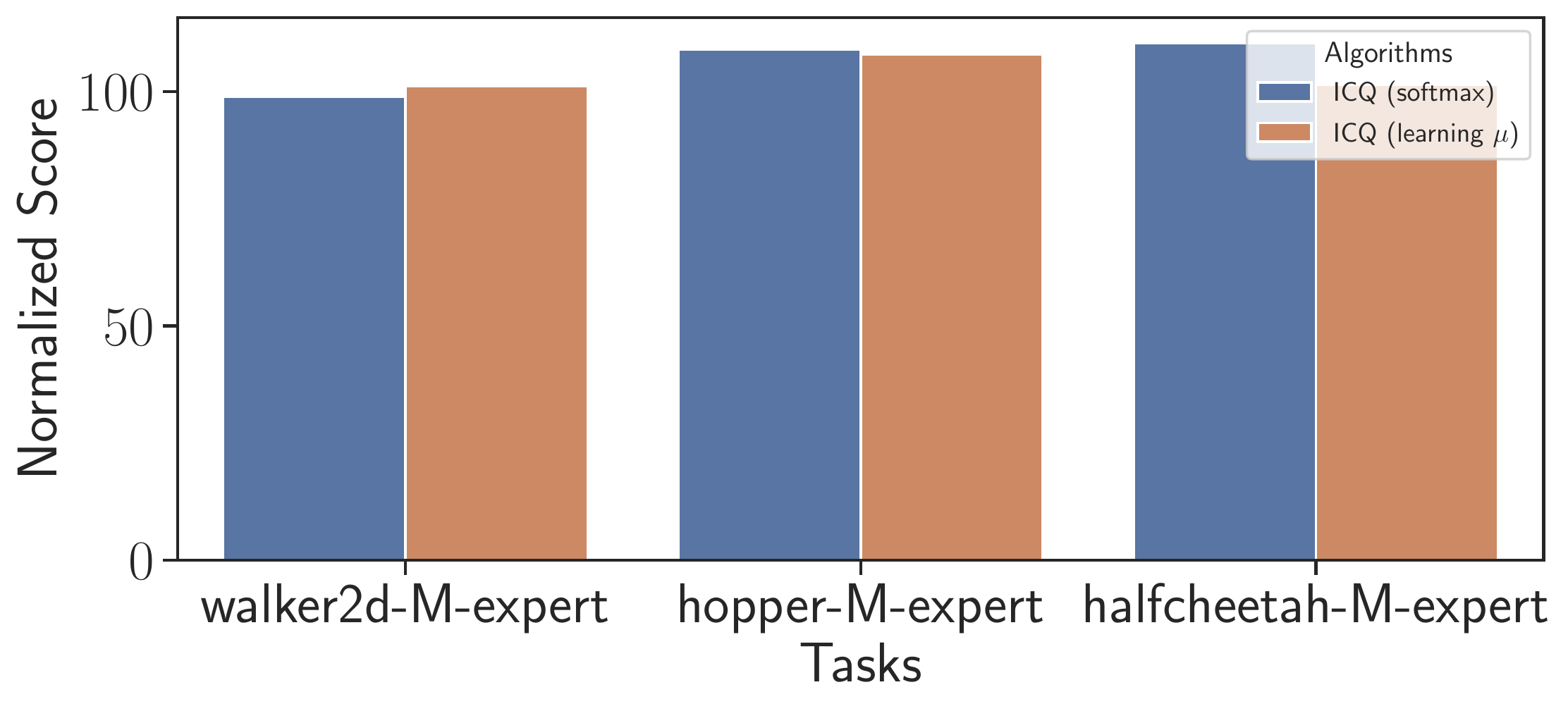}}
		\subfloat[Mujoco-medium.]
		{\includegraphics[width=2.3in]{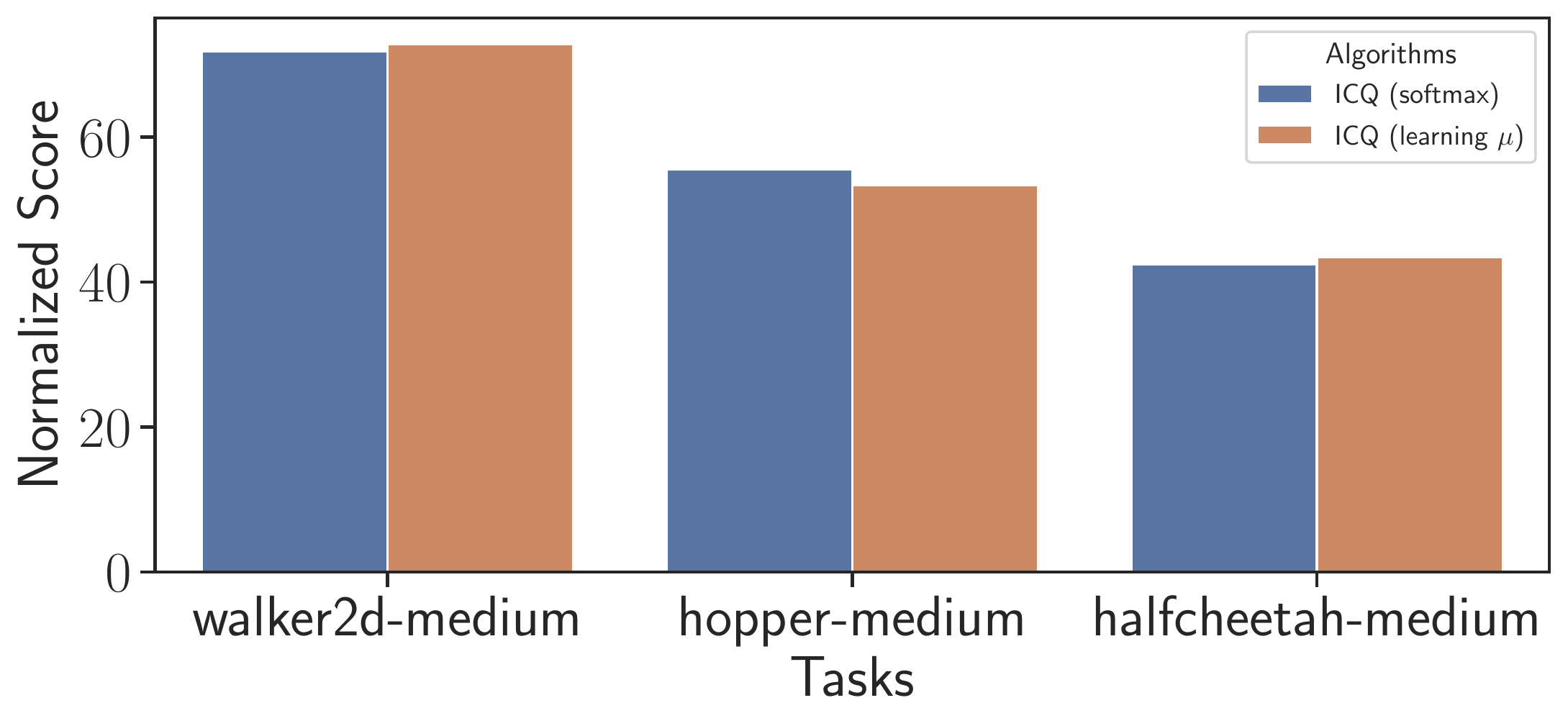}}
		\caption{The performance on D4RL tasks with different implementation of ICQ.}
		\label{different_implementation}
	\end{figure}

    \subsection{Ablation Study}\label{Appendx:ablation study}
	\begin{figure}[h]
		\centering
		%\subfloat[Parameter $\alpha$]
		{\includegraphics[width=2.35in]{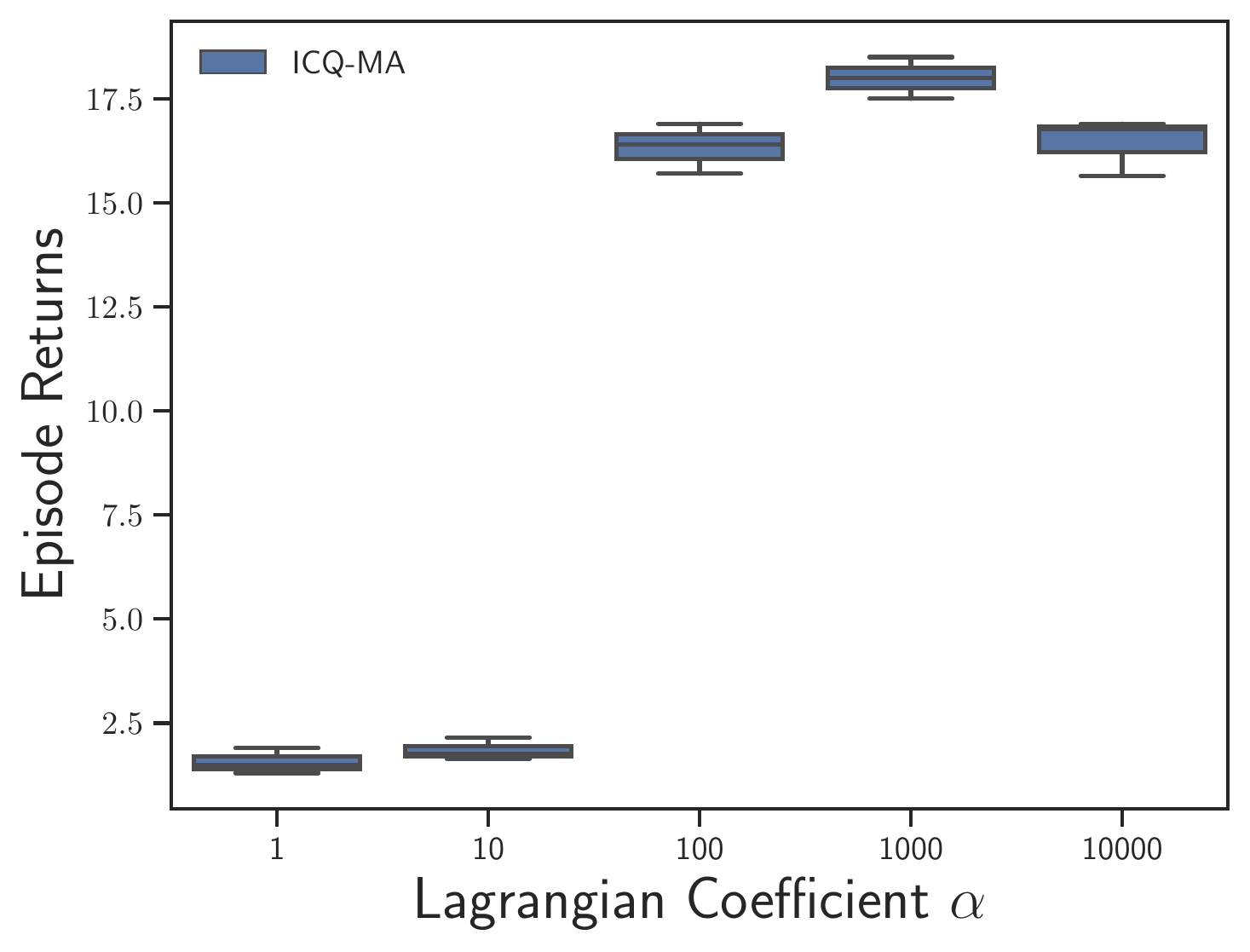}}
		\hspace{4.5mm}
		%\subfloat[Data quality.]
		{\includegraphics[width=2.3in]{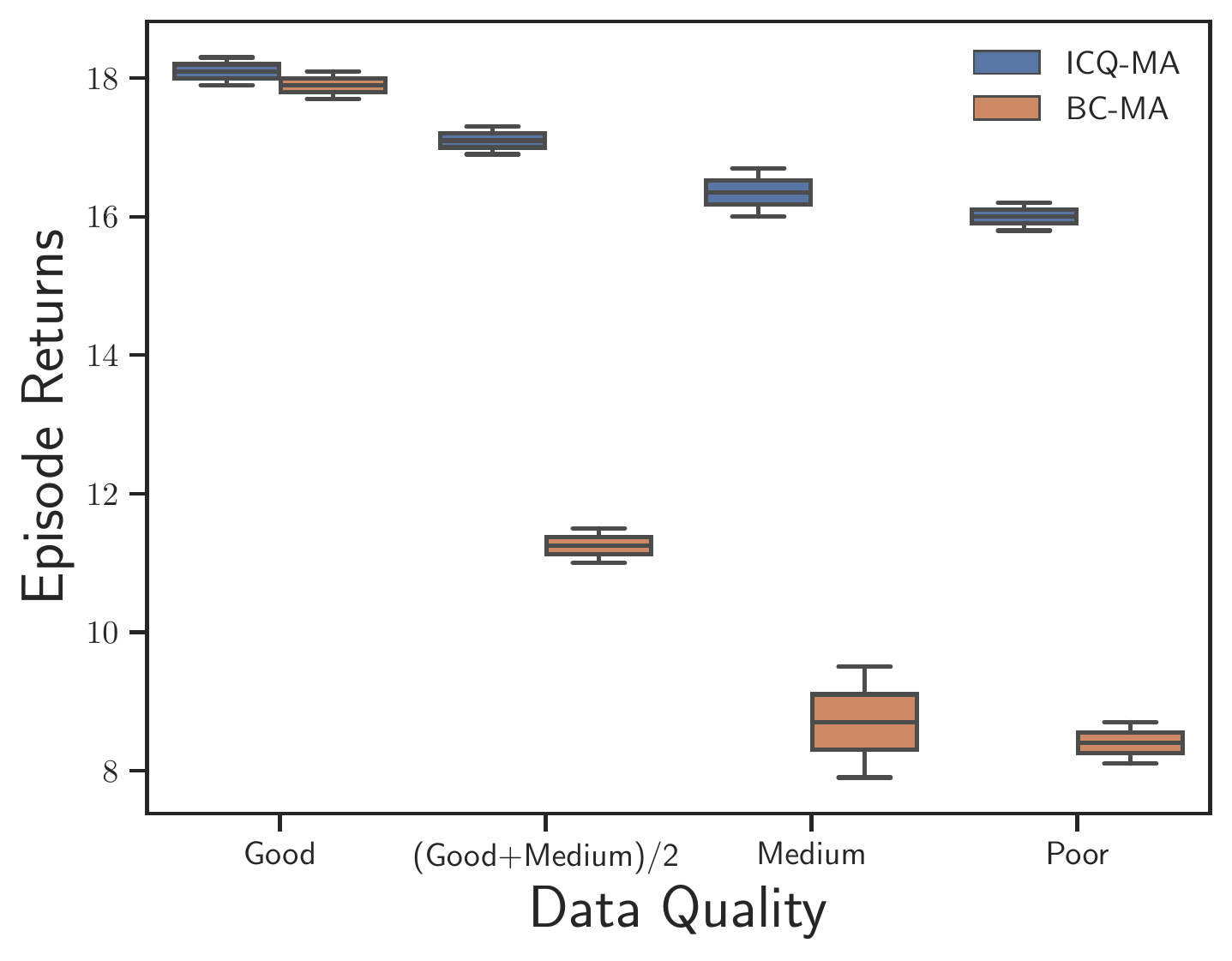}}
		\caption{Ablation study on MMM map.}
		\label{Ablation Study_2}
	\end{figure}

	\clearpage
	\section{Experimental Details}\label{experimental details}
	\subsection{Implementation details of ICQ}\label{ICQ Details}
	We provide the two implementation options of our methods regards whether learning $\mu$ to calculate $\rho$.
	
	\textbf{Learning an auxiliary behavior model $\hat{\mu}$.}
	We first consider to learn the behavior policy $\hat{\mu}$ using conditional variational auto-encoder as BCQ.
	Next, we will sample actions $n$ times ($n=100$ in our experiment) from $\hat{\mu}$ to calculate $Z(\tau)$ on each value update:
	\begin{equation}
	    \begin{aligned}
	    \rho(\tau, a) = \frac{\exp(\frac{Q(\tau,a)}{\alpha})}{Z(\tau)} \approx \frac{\exp(\frac{Q(\tau,a)}{\alpha})}{\mathbb{E}_{\tilde{a}\sim \hat{\mu}}\exp(\frac{Q(\tau,\tilde{a})}{\alpha})}.
	    \end{aligned}
	\end{equation}
	If $\hat{\mu} \approx \mu$, this method is favored as it provides an accurate approximation. However, since it may introduce unseen pairs sampled from the learned behavior model, it is against the principle of our analysis. Nevertheless, we believe it is still a better choice compared with BCQ. If there is any unseen pair $\tau, \tilde{a}$ with large extrapolation error sampled from $\hat{\mu}$, e.g, $Q_{\mathcal{B}}(\tau, \tilde{a}) \gg Q_{M}(\tau, \tilde{a})$, we will have $\hat{\rho}(\tau, a) < \rho(\tau, a)$, which means the unsafe estimation is truncated and the resulting target $Q$-value tends to be conservative.

	\textbf{Approximate with softmax operation over a mini-batch.}
	We have the following measure to approximately calculate $\rho$ without $\mu$, which reduces the computational complexity:
	
	\begin{equation}
	    \begin{aligned}
	    \rho(\tau, a) = \frac{\exp(\frac{Q(\tau,a)}{\alpha})}{Z(\tau)} \approx \frac{\exp(\frac{Q(\tau,a)}{\alpha})}{\sum_{(\tau',a')\sim \text{mini-batch}}\exp(\frac{Q(\tau^\prime, a^\prime)}{\alpha})},
	    \end{aligned}
	\end{equation}
	where $Z^i(\tau^i)$ is approximated by softmax operation over mini-batch samples. The benefit of the softmax operation is that it does not include any unseen pairs, which is consistent with our theoretical analysis. However, the price is that the softmax operation ignores the difference of states over a mini-batch, which introduces an additional bias. However, we find it does not harm the performance a lot in practice.
	There are also some previous works using softmax to deal with the partition function, such as AWAC~\cite{nair2020accelerating}) and VMPO~\cite{song2019v}, which has been confirmed to promote performance improvement.
	
	Considering the concise form of the softmax operation, we prefer the the second version in the multi-agent tasks.
	We conduct ablation studies of these two measures on D4RL to demonstrate their superior performance~(see Figure~\ref{different_implementation}).

	\subsection{Baselines Details}\label{Baselines Details}
	\textbf{BCQ-MA} is trained by minimizing the following loss:
	\begin{equation}
	\begin{aligned}
	\mathcal{J}^{\rm BCQ}_{Q}(\phi,\psi) & =\mathbb{E}_{\bm{\tau}\sim \mathcal{B}, \bm{a}\sim\bm{\mu}}\left[
	\left(r(\bm{\tau},\bm{a}) + \gamma \max_{\tilde{\bm{a}}^{[j]}}Q^{\bm{\pi}}(\bm{\tau}', \tilde{\bm{a}}^{[j]}; \phi', \psi') - Q^{\bm{\pi}}(\bm{\tau}, \bm{a}; \phi, \psi)
	\right)^2 \right]\\
	& \qquad \qquad \qquad \qquad \quad \tilde{\bm{a}}^{[j]} = \bm{a}^{[j]} + \xi(\bm{\tau}, \bm{a}^{[j]})
	\end{aligned},
	\end{equation}
	where $Q^{\bm{\pi}}(\bm{\tau}, \bm{a}) = w^i(\bm{\tau})Q^i(\tau^i,a^i) + b(\bm{\tau})$ and $\xi(\bm{\tau}, \bm{a}^{[j]})$ denotes the perturbation model, which is decomposed as $\xi^i(\tau^i, a^{i,[j]})$.
	If $\frac{a^{i, [j]} \sim G^i(\tau^i;\psi^i)}{\max\{a^{i, [j]} \sim G^i(\tau^i;\psi^i)\}_{j=1}^{m} } \leq \zeta$ in agent $i$, $a^{i, [j]}$ is considered an unfamiliar action and $\xi^i(\tau^i, a^{i,[j]})$ will mask $a^{i,[j]}$ in maximizing $Q^i$-value operation.
	
	\textbf{CQL-MA} is trained by minimizing the following loss:
	\begin{equation}
	\begin{aligned}
	\mathcal{J}_{Q}^{\rm CQL}(\phi, \psi) & = \alpha^{\rm CQL}\mathbb{E}_{\bm{\tau}\sim\mathcal{B}}\left[
	\sum_i \log\sum_{a^i}\exp(w^i(\bm{\tau})Q^i(\tau^i,a^i) + b(\bm{\tau})) - \mathbb{E}_{ \bm{a}\sim\bm{\mu}(\bm{a}\mid\bm{\tau})}[Q^{\bm{\pi}}(\bm{\tau}, \bm{a})]\right] \\
	& \qquad \qquad \qquad \qquad \qquad \qquad \qquad \qquad + \frac{1}{2}\mathbb{E}_{\bm{\tau}\sim\mathcal{B}, \bm{a}\sim\bm{\mu}(\bm{a}\mid\bm{\tau})}\left[
	\left(y^{\rm CQL} - Q^{\bm{\pi}}(\bm{\tau}, \bm{a})\right)^2\right] \\
	\mathcal{J}_{\bm{\pi}}^{\rm CQL}(\theta) & = \sum_{i}\mathbb{E}_{\tau^i,a^i\sim\mathcal{B}}\left[-\log(\pi^i(a^i\mid \tau^i;\theta_i)) Q^i(\tau^i, a^i)\right],
	\end{aligned}
	\end{equation}
	where we adopt the decomposed policy gradient to train $\bm{\pi}$, and $y^{\rm CQL}$ is calculated based on $n$-step off-policy estimation~(e.g., Tree Backup algorithm).
	Besides, $w^i(\bm{\tau}) = w^i(\bm{\tau};\psi)$, $b(\bm{\tau}) = b(\bm{\tau};\psi)$ and $Q^{\bm{\pi}}(\bm{\tau}, \bm{a}) = Q^{\bm{\pi}}(\bm{\tau}, \bm{a};\phi, \psi)$.
	
	\textbf{BC-MA} only optimize $\bm{\pi}$ by minimizing the following loss:
	\begin{equation}
	\begin{aligned}
	\mathcal{J}_{\bm{\pi}}^{\rm BC}(\theta) = \sum_i\mathbb{E}_{\tau^i, a^i\sim\mathcal{B}}[-\log(\pi^i(a^i\mid\tau^i;\theta_i))].
	\end{aligned}
	\end{equation}
	%%%%%%%%%%%%%%%%%%%%%%%%%%%%%%%%%%%%%%%%%%%
	
	%\clearpage
	\section{Multi-Agent Offline Dataset Based on StarCraft II}\label{Offline MARL Dataset}
	We divide maps in StarCraft II into three classifications based on difficulty~(see Table~\ref{Map in SMAC}).
	We divide behavior policies into three classifications based on the episode returns~(see Table~\ref{Behavior policy level}).
	\begin{table}[h]
		\caption{Classification of maps in the dataset.}
		\centering
		\begin{tabular}{cc}
			\toprule
			Difficulties & Maps \\
			\midrule
			Easy & MMM, 2s\_vs\_3z, 3s\_vs\_3z,
			3s5z, 2s3z, so\_many\_baneling \\
			Hard & 10m\_vs\_11m, 2c\_vs\_64zg \\
			Super Hard & MMM2, 27m\_vs\_30m \\
			\bottomrule
		\end{tabular}
		\label{Map in SMAC}
	\end{table}
	
	\begin{table}[h]
		\caption{Classification of behavior policies in the dataset.}
		\centering
		\begin{tabular}{cc}
			\toprule
			Level & Episode Returns \\
			\midrule
			Good & $15 \sim 20$ \\
			Medium & $10 \sim 15$ \\
			Poor & $0 \sim 10$ \\
			\bottomrule
		\end{tabular}
		\label{Behavior policy level}
	\end{table}
	%%%%%%%%%%%%%%%%%%%%%%%%%%%%%%%%%%%%%%%%%%%
	\subsection{Hyper-parameters}\label{hyperparameters}
	Hyper-parameters in multi-agent tasks are respectively presented in Table~\ref{Multi-agent hyper-parameters sheet}.
	Please refer to our official code for the hyper-parameter in single-agent tasks.
	
	\begin{table}[h]
		\caption{Multi-agent hyper-parameters sheet}
		\label{Multi-agent hyper-parameters sheet}
		\centering
		\begin{tabular}{ll}
			\toprule
			Hyper-parameter & Value \\
			\midrule
			Shared & \\
			\hspace{0.3cm} Policy network learning rate & $5 \times 10^{-4}$\\
			\hspace{0.3cm} Value network learning rate & $10^{-4}$\\
			\hspace{0.3cm} Optimizer & Adam \\
			\hspace{0.3cm} Discount factor $\gamma$ & 0.99 \\
			\hspace{0.3cm} Parameters update rate $d$ & 600 \\
			\hspace{0.3cm} Gradient clipping & 20 \\
			\hspace{0.3cm} Mixer network dimension & 32 \\
			\hspace{0.3cm} RNN hidden dimension & 64 \\
			\hspace{0.3cm} Activation function & ReLU \\
			\hspace{0.3cm} Batch size & 16 \\
			\hspace{0.3cm} Replay buffer size & $1.2\times 10^4$ \\
			\midrule
			Others & \\
			\hspace{0.3cm} Lagrangian coefficient $\alpha$ & 1000 or 100 \\
			\hspace{0.3cm} $\lambda$ & 0.8 \\
			\hspace{0.3cm} $\alpha^{\rm CQL}$ & 2.0 \\
			\hspace{0.3cm} $\zeta$ & 0.3 \\
			\bottomrule
		\end{tabular}
	\end{table}

\end{document}